\UseRawInputEncoding
\pdfoutput=1
\documentclass[preprint,authornum]{elsarticle}
\usepackage{amsthm, amssymb, amsmath}
\usepackage{hyperref}
\usepackage{multirow}%
\usepackage{pifont}
\usepackage{booktabs}%
\usepackage{graphicx}
\usepackage{subcaption}
\usepackage{pdflscape}
\usepackage{xspace}
\usepackage{breqn}
\usepackage{algorithm}
\usepackage{algpseudocode}
\usepackage{colortbl} 
\usepackage{rotating}
\usepackage{orcidlink}
\usepackage{cleveref}

\theoremstyle{definition}%

\newcommand{\Ph}{\mathbb{P}}

\newcommand{\DD}{\mathcal{D}}

\newcommand{\RR}{\mathcal{R}}

\newcommand{\cc}{\mathbf{c}}

\newcommand{\ee}{\mathbf{e}}

\newcommand{\uu}{\mathbf{u}}
\newcommand{\vv}{\mathbf{v}}

\newcommand{\xx}{\mathbf{x}}
\newcommand{\yy}{\mathbf{y}}

\newcommand{\FB}[1]{{\color{olive} FB: #1}}

\newcommand{\goproo}{\texttt{GoProO}\xspace}
\newcommand{\gopros}{\texttt{GoProS}\xspace}
\newcommand{\goprors}{\texttt{GoProRS}\xspace}
\newcommand{\bsd}{\texttt{BSD}\xspace}

\newcommand{\speinet}{\texttt{SPEINet}\xspace}
\newcommand{\vrt}{\texttt{VRT}\xspace}
\newcommand{\vdtr}{\texttt{VDTR}\xspace}
\newcommand{\cdvd}{\texttt{TSP}\xspace}
\newcommand{\cdvdnl}{\texttt{TSPNL}\xspace}
\newcommand{\dstnet}{\texttt{DSTNet}\xspace}

\newcommand{\pvd}{\texttt{PVDNet}\xspace}

\newcommand{\dtwonet}{\texttt{D\textsuperscript{2}Net}\xspace}

\journal{Pattern Recognition}
\begin{document}
\begin{frontmatter}




\title{Video Deblurring by Sharpness Prior Detection and Edge Information}

\author[label1,label3]{Yang Tian\orcidlink{0000-0002-2882-973X}}
\ead{tianyang@hrbeu.edu.cn}

\author[label2,label3]{Fabio Brau\orcidlink{0000-0002-9948-111X}}
\ead{fabio.brau@unica.it}
\author[label3]{Giulio Rossolini\orcidlink{0000-0002-6404-2627}}
\ead{g.rossolini@sssup.it}

\author[label3]{Giorgio Buttazzo\orcidlink{0000-0003-4959-4017}}
\ead{g.buttazzo@sssup.it}
\author[label1]{Hao Meng\orcidlink{0000-0003-3586-9286}}
\ead{menghaoHEU@hotmail.com}

\affiliation[label1]{organization={The College of Intelligent Systems Science and Engineering, Harbin Engineering University}, 
             addressline={Nantong, 145}, 
             city={Harbin}, 
             postcode={150001}, 
             country={China}}

\affiliation[label2]{organization={DIEE, University of Cagliari},
             addressline={Via Università, 40}, 
             city={Cagliari}, 
             postcode={09124}, 
             state={Italy}}

\affiliation[label3]{organization={Dept. of Excellence in Robotics and AI, Sant'Anna School of Advanced Study},
             addressline={Via Moruzzi, 1}, 
             city={Pisa}, 
             postcode={56124}, 
             state={Italy}}
\begin{abstract}
Video deblurring is essential task for autonomous driving, facial recognition, and security surveillance. Traditional methods directly estimate motion blur kernels, often introducing artifacts and leading to poor results. Recent approaches utilize the detection of sharp frames within video sequences to enhance deblurring. However, existing datasets rely on fixed number of sharp frames, which may be too restrictive for some applications and may introduce a bias during model training. To address these limitations and enhance domain adaptability, this work first introduces GoPro Random Sharp (\goprors), a new dataset where the the frequency of sharp frames within the sequence is customizable, allowing more diverse training and testing scenarios. Furthermore, it presents a novel video deblurring model, called \speinet, that integrates sharp frame features into blurry frame reconstruction through an attention-based encoder-decoder architecture, a lightweight yet robust sharp frame detection and an edge extraction phase. Extensive experimental results demonstrate that \speinet outperforms state-of-the-art methods across multiple datasets, achieving an average of +3.2\% PSNR improvement over recent techniques. 
Given such promising results, we believe that both the proposed model and dataset pave the way for future advancements in video deblurring based on the detection of sharp frames.
\end{abstract}
\end{frontmatter}
\section{Introduction}


With the increasing development of technology and social media, video recording using handheld and airborne devices such as smartphones, action cameras and drones is becoming increasingly popular \citep{xian2024mitfas}. However, fast light changes, camera shakes, rapid object movements and depth changes can often produce undesired effects and lead to video blur \citep{rota2023video}. The presence of blur in a video not only affects the visual quality, but can also hinder other downstream visual tasks such as tracking \citep{sun2024long} and video anomaly detection \citep{nayak2021comprehensive}. While dedicated hardware, such as event cameras, could be beneficial \citep{chakravarthi2024recent}, it represents a costly and partial solution that is not always applicable to popular devices.

For this reason, the development of effective video deblurring algorithms has been actively studied in the last decades \citep{kim2015generalized} and still represents a hot research topic \citep{imani2024stereoscopic}.
To recover sharp frames from blurred video, conventional approaches \citep{kim2015generalized,delbracio2015handheld,pan2017simultaneous,ren2017videodeblurring} typically use optical flow to estimate a blur kernel and recover the blur sequence by deconvolution, making assumptions about the type of motion blur. However, the dynamic nature of video and the difficulty of estimating the motion blur kernel often leads to artifacts and poor robustness.

With the advancement of artificial intelligence, video deblurring methods based on deep learning have achieved remarkable success \citep{ren2024fast,song2024memory,cheng2022video,amini2021medical,imani2024stereoscopic,cao2022vdtr}. To achieve better performance, other works \citep{zhu2022dastnet,lin2024lightvid,pan2023deep,suin2021gated} focused on the spatio-temporal dependence of successive non-uniformly blurred video frames. For example, \citet{wang2021video} 
proposed the Spatio-Temporal Pyramid Network (SPN), which can dynamically learn different spatio-temporal dependencies, while \citet{zhu2023exploring} used time-spectrum energy as a global clarity guide to solve the problem of local feature loss due to blur.
Furthermore, \citet{lin2020learning} and \citet{kim2022event} incorporate the active gray level pixels generated by an event camera to further improve the quality of the results. 


Recently, some research has observed that not all frames in a blurred video are actually blurred, and that some \textit{sharp frames} may occasionally occur, \citep{pan2023cascaded, xiang2020deep, shang2021bringing}. 
Building on this observation, previous studies have suggested that once these sharp frames are identified, they can be exploited to extract finer and more precise temporal information, enhancing the deblurring process.
For example, \citet{pan2023cascaded} proposed a convolutional neural network (CNN) that incorporates a temporal sharpness prior module to utilize the information from sharp frames for recovering blurred frames. Similarly, \citet{xiang2020deep} leveraged features from sharp frames to estimate temporal information related to video motion flow, guiding the network to restore blurred images.

In addition to the abovementioned techniques, new ad hoc datasets that incorporate portions of sharp frames have gained increasing attention, such as the GoProS dataset \citep{shang2021bringing}. However, in these datasets, sharp frames are added to the sequences based on a probability that is consistently constrained to 50\%, resulting in a fixed ratio of sharp to blurred frames.
This constraint makes such datasets unsuitable for videos where the occurrence of sharp and blurry frames varies dynamically, significantly limiting their applicability in real-world scenarios.

To overcome these problems, this paper introduces a new dataset, \goprors, in which sharp frames occur according to a probability defined by the user, so enabling a desired ratios of sharp and blurred frames. This setup helps trained models to generalize across diverse blurring conditions.
Then, to evaluate the benefits of the dataset and advance the state-of-the-art in deblurring, this paper proposes a novel video deblurring method called Sharpness Prior Detection and Edge Information Network (\speinet). 
The \speinet model incorporates regularized edge emphasis to improve the edge mapping capabilities of blurred frames and includes a lightweight sharp frame detector based on six computationally efficient classical metrics commonly used for autofocus \citep{pertuz2013analysis}, avoiding reliance on a slower, dedicated neural network as used in previous works \citep{shang2021bringing}. Finally, its attention-based encoder-decoder architecture effectively integrates sharp frame information and edge-enhanced features to reconstruct blurred video frames. In particular, the encoder extracts deep features from sharp and blurry frames using attention mechanisms, while the self-transfer and cross-transfer modules transfer key pixel details to assist the decoder in recovering and reconstructing blurred frames.


In summary, the paper provides the following main contributions:
\begin{enumerate}
\item A new video deblurring dataset (\goprors) is proposed and synthesized, where sharp frames occur with a probability defined by the user, thus enhancing the model's generalization capabilities.
\item The \speinet model for video deblurring is proposed, which leverages key pixel information from sharp frames to guide the recovery and reconstruction of adjacent blurry frames. The model incorporates an edge emphasis algorithm and a lightweight sharp frame detector, integrated with a custom designed attention-based encoder-decoder architecture. 

\item Extensive experiments have been carried out to evaluate the performance of \speinet on four datasets, demonstrating the effectiveness of the proposed approach and the benefits of the \goprors dataset. The achieved results show that \speinet outperforms the state-of-the-art methods across multiple datasets, providing a clear direction for future research and advancements in video deblurring.

\end{enumerate}

The rest of the paper is organized as follows. \Cref{sec:relworks} analyzes the state of the art, focusing on the solutions that use sharp frames to improve the results. \Cref{sec:dataset} presents the construction details of the \goprors dataset.
\Cref{sec:speinet} presents the framework and components of the model. \Cref{sec:experiments} presents the experimental results comparing the performance of \speinet with the ones of other related state-of-the-art models. \Cref{sec:ablation} includes a detailed ablation study to assess the impact of each component of the model. \Cref{sec:visualization} illustrates some examples of deblurred video sequences and, finally, \Cref{sec:conclusion} concludes the paper.
\section{Related works}\label{sec:relworks}

With the invention of cameras and imaging devices, the demand for image deblurring has increased. A large plethora of methods for image deblurring (e.g., \cite{su2017deep}) have been proposed in the literature, and a great part is based on residuals networks, transformers, and diffusion models, that recently have achieved state-of-the-art performances \citep{zhang2022deep}. For instance, the hierarchical integrated diffusion model proposed by \cite{chen2024hierarchical} achieved a realistic image deblurring, while
\citet{ren2023multiscale} introduced a simple but effective multi-scale structure guidance as an implicit bias to improve the deblurring results. However, diffusion models require many inference iterations, resulting in a large computational resource consumption. 

Recently, attention-based global and local end-to-end CNN models have also been proposed. For example, \citet{mao2023intriguing} inserted kernel-level information into a convolutional network for image deblurring, while \citet{yan2023sharpformer} proposed to directly learn long-range dependencies to overcome large non-uniform blur variations. \citet{huang2024effective} proposed a blurred attention block and skip connection into sub-networks to restore the edges and details of the object. \citet{cui2023dual} exploited self-attention obtained through a dynamic combination of convolutions and integrated local feature information in the encoding stage. \citet{dong2023multi} combined low-pass filters and wavelet feature fusion to explore more image details. \citet{kong2023efficient} developed an efficient frequency-domain-based self-attention solver that estimates scaled dot-product attention by element-wise product operations instead of matrix multiplication in the spatial domain to restore image details.


In contrast to single image deblurring, video deblurring methods can exploit additional information between adjacent images. Video deblurring not only requires restoring the clarity of a single frame, but also requires ensuring the temporal consistency between multiple frames to prevent flickering or artifacts during video playback. Therefore, video deblurring faces many challenges.
Early video deblurring methods usually involve object motion estimation \cite{kim2015generalized, delbracio2015handheld, pan2017simultaneous,ren2017videodeblurring}. \citet{kim2015generalized} use bidirectional optical flow to approximate pixel-level kernels while estimating optical flow and latent frames to achieve video deblurring. Specifically, \citet{ren2017videodeblurring} use different motion models for image regions to guide optical flow estimation to achieve video deblurring, while other works, e.g., \citet{delbracio2015handheld}, eliminate blur caused by camera shake by combining Fourier domain information from nearby frames in the video. Although these methods produce valuable deblurring results, they rely on assumptions that often do not generalize to different scenarios and are unsuitable for videos containing sudden scene changes.

With the advancement of vision transformers, spatial attention mechan ism has increasingly been adopted in video deblurring models \citep{ren2024fast, imani2024stereoscopic, cao2022vdtr}. For instance, \citet{ren2024fast} designed an asymmetric encoder-decoder architecture with residual channel spatial attention blocks to improve the performance, while \citet{imani2024stereoscopic} and \citet{cao2022vdtr} leveraged a self-attention mechanism for increasing the deblurring quality through image-feature alignment.
To further improve performance, \citet{zhu2022dastnet} considered spatio-temporal information and proposed a deep perception modulation block to capture spatio-temporal features from the depth map, while \citet{pan2023deep} dynamically adjusted the importance of spatial and temporal information. 
\citet{weiaggregating2025} exploits hybrid transformation to aggregate information from sharp frames to blurry neighboring frames. \citet{lin2024lightvid} leveraged a hierarchical
spatial feature extraction to separate the blurry region, achieving lightweight improvements. Finally, \citet{suin2021gated} focused on better integrating and collecting spatio-temporal information in blurry video frames. \citet{rao2025rethinking} exploits the wavelet deal with the high-frequency information generate by diffusion model. \citet{he2025domain} considered a new training scheme to calibrate the target domain of the model in the test and solve the problem of domain adaptation difficulty of the video deblurring model.

Along with the development of new models, some research also noticed that the deblurring performance can be improved by exploiting not only the dependence between adjacent frames, but also the presence of sharp frames. This also help reducing the introduction of artifacts in the output. For this reason, a growing number of researches started tackling the deblurring problem with the information retained in sharp frames that naturally occur in a blurred videos, through detection or attention mechanisms.

For example, \citet{pan2020cascaded} designed a CNN model to estimate optical flow from intermediate sharp frames and then used the estimated optical flow to recover the blurred frames. \citet{shen2021spatial} designed an enhanced optical flow by exploiting complementary information from short-exposure and long-exposure inputs. 
\citet{shang2021bringing} introduced the \gopros dataset to train the \dtwonet model, which uses the position of sharp frames inferred by a \texttt{Bi-LSTM} sharp detector to recover the blurred frames.
\citet{zhu2023exploring} proposed a spatio-temporal clarity map that implicitly uses generative networks to learn inter- and intra-frame sharp priors, while \citet{zhang2024blur} exploited a blur map that transforms the original dense attention mechanism into a sparse form to better transfer the information from sharp frames to the blurred adjacent ones. Finally, \citet{song2024memory} proposed a memory gradient-guided progressive propagation network in which the memory branch stores the blurred-sharp feature pairs, providing valuable information for the deblurring process.

However, it is worth noting that all the algorithms discussed above are designed to deal with videos that contain a fixed, predetermined number of sharp frames. However, in real-world scenarios, the number of blurred and sharp frames can vary unpredictably due to sudden changes in the scene, camera movements, and exposure time of the camera, resulting in poor domain adaptation of the model. Therefore, it is very important to work on video deblurring in scenarios with unfixed blurry and sharp frames.

\section{GoProRS dataset}
\label{sec:dataset}
In the following, we propose the GoPro Random Sharp (\goprors) dataset to improve the evaluation and training of deblurring models in real-world blurry scenes. The main goal of the \goprors dataset is to introduce a variable ratio of sharp frames available in the sequence, addressing a wide variety of blurring scenarios and thus enhancing the generalization capability.

\goprors is based on the GoPro dataset proposed by \cite{nah2017deep}, which contains $33$ blurred video sequences generated by averaging consecutive frames obtained from a high-frequency camera. We will refer to this dataset as GoPro Original (\goproo). Specifically, in \goproo , each frame in the blurred sequence is created by averaging the frames within a window of length $w$ variable in the range $\{7, \dots, 13\}$,
where the central frame in the window is used as the ground truth.
Note that $w$ is randomly chosen for each frame, and all the resulting frames are visibly blurred. 

Formally, \goprors is a set of $N_v=33$ triplets $\DD_r:=\{(x^{(i)},l^{(i)},g^{(i)})\}_{i=1:N_v}$ where $x^{(i)}$ is the blurred video sequence, $l^{(i)}$ is a vector of binary labels that indicates for each $j$ if the frame $x_j^{(i)}$ is sharp/blurred, and $g^{(i)}$ is the ground-truth sequence. Let $\RR=\{v^{(i)}\}_{i=1:N_v}$ be the high-frame-rate (HFR) videos used for crafting \goproo, containing $N_v$ RGB sharp videos of shape $H\times W = 1280\times 720$ captured at $250$ $fps$. For a given ratio $r\in[0,\frac12]$,  and for each video-sequence $v^{(i)}\in[0,1]^{L_i\times3\times H \times W}$ a vector of window lengths $w^{(i)}\in\{1,\ldots,15\}^{N_i}$ is sampled in such a way that $\Ph(w_j^{(i)} \le 5)=r$ and $N_i = \sum_j w_j^{(i)}\le L_i$. Then, the frames are aggregated by averaging, without overlaps, considering the sampled window lengths. In formulas, the triplets are defined by
\begin{equation}
    x_j^{(i)}=\frac{1}{w_j^{(i)}}\sum_{k=0}^{w_j^{(i)}-1} v_j^{(\sigma_j^{(i)}+k)},\quad l_j^{(i)}=(w_j^{(i)}\le5),\quad g_j^{(i)} = x_i^{(\sigma_j^{(i)}+w_j^{(i)}//2)},
\end{equation}
for each $j<N_i$, where $\sigma_j^{(i)}=\sum_{k< j} w_j^{(i)}$. Note that this procedure guarantees that, on average, the proportion of sharp frames in each video is close to $r$. \Cref{tab:ratios} contains the measured ratios for each video.


It is important to remark that the proposed \goprors dataset is a natural generalization of \gopros \cite{shang2021bringing}. In fact, while the latter constrains the ratio between sharp and blurred frames to be 0.5, \goprors, instead, allows the user to set the sharp/blur ratio $r$ between $0$ and $0.5$. 
There are two reasons for this choice. First, we believe this dataset is more aligned with real-world scenarios, where blurred frames may occur due to sudden movement of the camera or the subject in the video. Second, as clearly stated in \Cref{sec:comparisons}, the \dtwonet model, trained on \gopros, shows a significant drop in performance when tested on the \goproo dataset. This indicates that training on \gopros cannot generalize well to more heavily blurred datasets like \goproo. This highlights the need for a dataset containing video sequences with a wider range of sharp-frame ratios.

\section{The \speinet framework}\label{sec:speinet}
This section introduces the \speinet framework, detailing its processing flow and each of its functional modules.

\begin{figure*}[!ht]
\centering
\includegraphics[width=\linewidth]{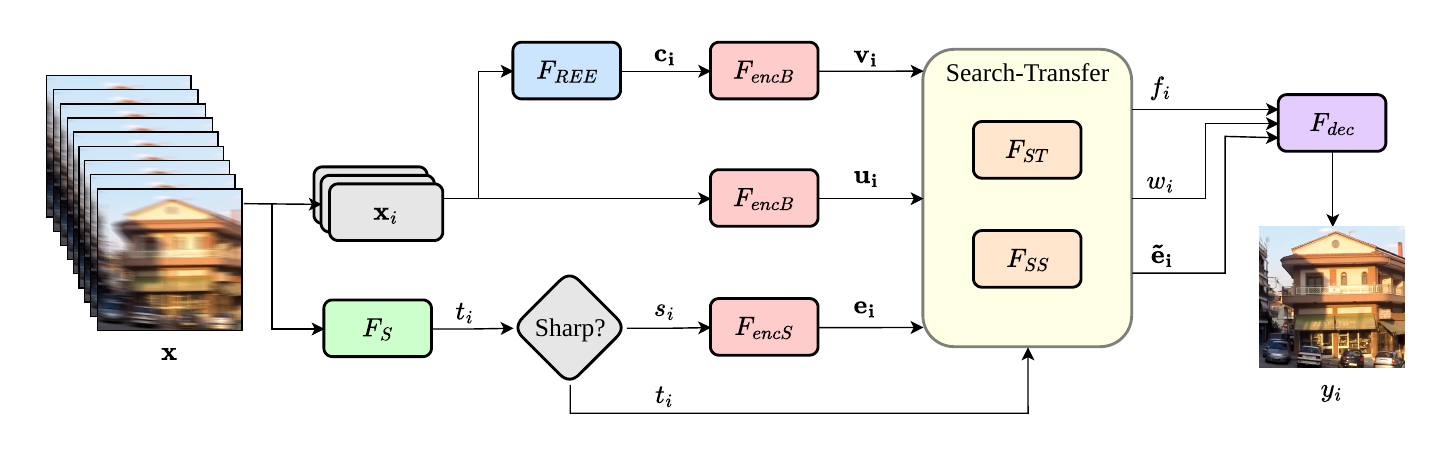}
\caption{Overview the SPEINet framework during the inference of the model. $F_{S}$ represents the sharp detection of the Stage 1. $F_{REE}$ represents the edge extraction of Stage 2; $F_{encB}$ and $F_{encS}$ represent the encoding performed at Stage 3. The modules in Search-Transfer represent the Stage 4. The $F_{dec}$ performs the reconstruction of Stage 5.}
\label{fig:framework}
\end{figure*}

\subsection{Overview of the model}
The \speinet framework processes video sequences by breaking them into overlapping triplets and reconstructing each frame using sharp frame detection, edge extraction, feature encoding, information transfer, and decoding.
In the following, $\xx$ denotes an input video sequence, while $\xx_{i:j}$ denotes a sub-sequence from the $i$-th to the $j$-th frames included. Also, $x_i$ indicates the $i$-th frame and $\xx_i \!:=\xx_{i-1:i+1}$ identifies the triplet of frames centered in $x_i$. The frames in $\xx$ are processed by the architecture summarized in \Cref{fig:framework}. The architecture includes five stages that interoperate to produce the unblurred frame sequence $\yy$. The proposed pipeline is inspired by \dtwonet \citep{shang2021bringing}, which is also based on sharp frames, but it is redefined to better generalize on more dynamic blur sequences, as demonstrated in the experimental results on the proposed \goprors dataset. Each stage of the pipeline is  briefly described below.

\begin{itemize}
\item \textit{Stage 1 (Sharp detection)} The sharp-detection module $F_S$ finds, for each frame-index $i$, the index $t_i=F_S(\xx,i)$ of the closest sharp frame to $x_i$ among the seven past frames. If it is found, it is denotes as $s_i=x_{t_i}$, otherwise we set $t_i=-1$.

\item \textit{Stage 2 (Edge extraction)} For each frame $i$, function $F_{\text{REE}}$ extracts edge information producing a new frame \(c_i = F_{\text{REE}}(x_i)\) by applying a Regularized Edge-Emphasizing through the Richardson-Lucy method \citep{richardson1972bayesian}.
For consistency, $\cc_i$ denotes the triplet of frames with emphasized edges obtained by applying $F_{\text{REE}}$ to $\xx_i$ frame-wise.

\item \textit{Stage 3 (Encoding)} For each $i$, the blur-encoder $F_{\text{encB}}$ encodes the triplets $\xx_i$ and $\cc_i$ to produce the triplets of features $\uu_i$ and $\vv_i$, respectively.
If the closest sharp frame $s_i$ exists, the sharp-encoder $F_{\text{encS}}$ encodes the sharp frame $s_i$ to produce a triplet of features $\ee_i=(e_i^{(1)},e_i^{(2)},e_i^{(3)})$ with the cascade approach described in \Cref{sec:encoding}.

\item \textit{Stage 4 (Search and Transfer)} 
If the sharp frames $s_i$ exists, the \textit{search-and-transfer} module $F_{\text{ST}}$ combines the information retained in the previously computed triplets $\uu_i,\,\vv_i,\,\ee_i$ to deduce the intermediate latent triplet $\tilde\ee_i$, and the latent frames $f_i$, $w_i$. 
Otherwise, the \textit{self-search} module $F_{\text{SS}}$ only process the information of the encoded blurred frames. In formulas, 
\begin{equation}
(f_i,w_i,\tilde \ee_i) =
\begin{cases}
    F_{\text{SS}} (\uu_i, \vv_i) & \text{if } t_i = -1 \\
    F_{\text{ST}} (\uu_i, \vv_i, \ee_i) & \text{otherwise} 
    \end{cases}.
\label{eq:search_and_transfer}
\end{equation}

\item \textit{Stage 5 (Decoding)} The decoding module $F_{dec}$ decodes the latent features $f_i$, $\tilde\ee_i$, and $w_i$, to deduce the un-blurred frame $y_i = F_{\text{dec}}(f_i, w_i,\tilde\ee_i)$, by leveraging the multi-scale operation described in the dedicated section.
\end{itemize}

An algorithmic description of the framework is also given in \Cref{alg:speinet}, whereas a detailed explanation of the five sub-modules is provided below.

\begin{algorithm}
\scriptsize
\caption{\speinet pseudocode}
\label{alg:speinet}
\begin{algorithmic}[1]
\State \textbf{Input:} Blurry video sequence $\xx$
\State \textbf{Output:} Restored video sequence $\yy$
\State $N \gets \texttt{Length}(\xx)$
\For{$i = 1 \to N - 1$}
    \State $t_i \gets F_S(\xx, i)$\Comment{Sharp detection Stage}
    \State $\xx_i \gets (x_{i-1}, x_i, x_{i+1})$ 
    \State $\cc_i \gets F_{\text{REE}}(\xx_i)$ \Comment{REE Stage}

    \State $\uu_i \gets F_{\text{encB}}(\xx_i)$\Comment{Blur-encoding stage}
    \State $\vv_i \gets F_{\text{encB}}(\cc_i)$

%
    \If{$t_i \neq -1$} \Comment{Search-Transfer Stage}
        \State $s_i \gets x_{t_i}$
        \State $\ee_i \gets F_{\text{encS}}(s_i)$
        \State $f_i, w_i, \tilde{\ee}_i \gets F_{\text{ST}}(\uu_i, \vv_i, \ee_i)$
    \Else \Comment{Self-Search Stage}
        \State $f_i, w_i, \tilde{\ee}_i \gets F_{\text{SS}}(\uu_i, \vv_i)$
    \EndIf
    \State $y_i \gets F_{\text{dec}}(f_i, w_i, \tilde{\ee}_i)$
\EndFor
\State \Return $\{y_i\}_{i=1}^{N-1}$
\end{algorithmic}
\end{algorithm}

\subsubsection{Sharpness prior detection}
\label{sec:sharp-frame-detector}

The sharp frames contained in the video can be leveraged to extract detailed information useful to improve the deblurring process. To this end, it is crucial to implement an efficient method that detects them.

In \dtwonet, deblurring is performed by leveraging the left and right sharp frames closest to $x_i$ \citep{shang2021bringing}. Such sharp frames are detected by training a \texttt{Bi-LSTM} sharp-binary detection model to the video sequence $\xx$. 
In the proposed \speinet model, instead, we adopt a more classical approach that provides better generalization in detecting sharp frames, even when their availability varies within video sequences. 
Specifically, only a single sharp frame \( s_i \) closest to \( x_i \) is searched within a window of \( \gamma \) past frames, where \( \gamma = 7 \) was used in our experiments. We found that a larger value of $\gamma$ is not advisable, as sharp frames outside this range may differ significantly from the blurred frame \( x_i \). 
If no sharp frame is found within this window, a switching mechanism explained in \Cref{sec:self-search} is applied.



\label{sec:sharp-frame-detection}
\begin{figure}[h]
    \centering
\includegraphics[width=0.7\linewidth]{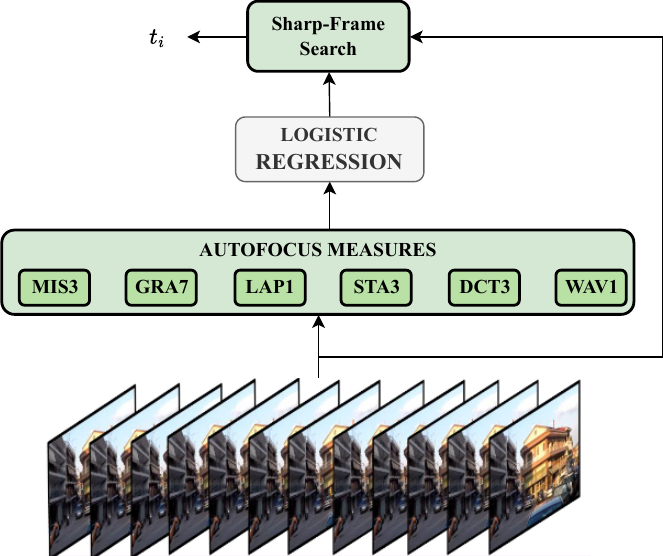}
    \caption{Schema of Sharp Frames Detection. Logistic regression is trained to perform binary classification on each frame that can be \texttt{sharp/blur}. The $F_{S}$ searches for the closest sharp frame to $x_i$.}
    \label{fig:goprors}
\end{figure}

As shown in \Cref{fig:goprors}, the detection of blurred and sharp frames is treated as a binary classification task, i.e., distinguishing and detecting sharp frame coordinates from blurred frame coordinates. To avoid the additional computational cost of a dedicated neural network and the risk of getting into overfitting scenarios, such sharp frames detection is performed by leveraging a logistic-regression algorithm applied to six metrics originally devised for automatic focus by \cite{pertuz2013analysis}.

More specifically, given a single frame $x_i$, the following six auto-focus metrics are computed: \texttt{MIS3, GRA7, LAP1, STA3, DCT3, WAV1} (a detailed description of such metrics can be found in \ref{app:autofocus}). These six scalar values are concatenated in a vector and then classified through \textit{Logistic Regression} to deduce a binary \texttt{sharp/blur} label $\hat l_i$. Note that the logistic-regression model is fastly trained in a supervised manner on ground truth labels. Given a video $\xx$ and a frame index $i$, the module $F_S$ returns an index $t_i$ by first classifying all the frames in the window $x_{i-\gamma:i}$ and looking for the closest sharp frame to $x_i$. As anticipated above, if all the frames in $x_{i-\gamma:i}$ are classified as blur, then $t_i$ is set to $-1$ in order to trigger a self-search stage in the subsequent processing part.

Since different classical detection strategies can be used, we compared the Logistic Regression (LR) with different standard classification methods, such as Decision Tree (DT) and Random Forest (RF), see \ref{sec:abl-kernel} for further details. Furthermore, as it will be discussed in \Cref{sec:abl-detection}, this method introduces a negligible overhead with respect to a \texttt{Bi-LSTM} neural model without degradating the performance of \speinet. 

\subsubsection{Regularized edge emphasizing}
The edges of objects in non-uniform blurred frames are smooth, and the edge information is lost. Therefore, a restoration of edge information can effectively improve the performance of the model deblurring and is crucial for video deblurring. 
Traditional edge highlighting algorithms mostly use Sobel \citep{sobel19683x3} and Laplace \citep{marr1980theory} operators to restore edge information with low computational effort. However, the edge-sharpening obtained by these techniques is not the optimal one even in presence of uniform-blur. 
The Richardson Lucy (RL) method instead \citep{richardson1972bayesian} applies an optimal deconvolution to the uniform blur kernel, producing an effective image sharpening, allowing a fast restoration of edges. 
For these reasons, our model integrates a \textit{Regularized Edge Emphasis} phase based on the RL method for sharpening the image, highlighting the edges, 
thus enabling a better processing by the subsequent modules, as shown in the ablation studies reported in Section \ref{ss:ablation_submodule}.
\begin{figure}[h!]
    \centering
    \begin{subfigure}[b]{0.7\columnwidth}
        \centering
        \includegraphics[width=\columnwidth, clip, trim=0 0.28cm 0 0.27cm]{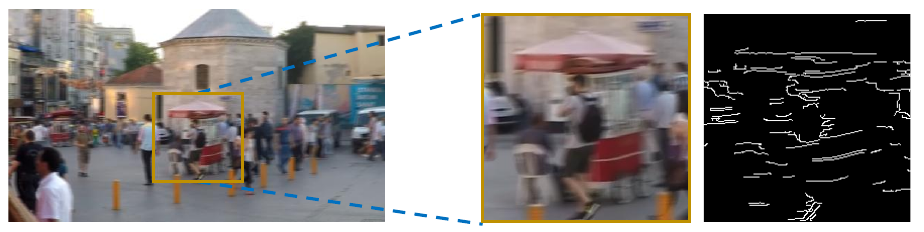}
        \caption{Blurred input frame}
        \label{fig:subfig1}
    \end{subfigure}
    
    \begin{subfigure}[b]{0.7\columnwidth}
        \centering
        \includegraphics[width=\columnwidth, clip, trim=0 0.28cm 0 0.35cm]{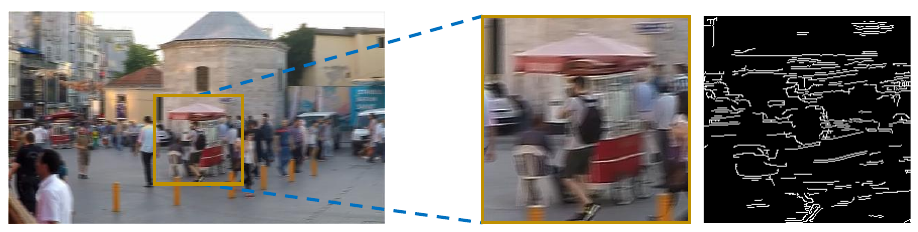}
        \caption{Frame with emphasized edges}
        \label{fig:subfig2}
    \end{subfigure}    
    \caption{Edge extraction obtained by applying the regularized-edge-emphasizing module $F_{\text{REE}}$ to the frame in \Cref{fig:subfig1}.}
    \label{fig:edge-emphasizing}
\end{figure}

\Cref{fig:edge-emphasizing} shows the effect of the algorithm on enhancing the edges of blurred frames. We can observe that the blurred image~\ref{fig:subfig1} has smooth contours and loses a lot of edge detail information, while the contour edges in the image~\ref{fig:subfig2} generated by the regularized edge emphasis module are enhanced.
At this stage, the algorithm processes each frame $x_i$ and returns a frame $c_i$. We also use the symbol $\cc_i$ to represent the triple centered at $c_i$.

\subsubsection{Attention-based encoder}
\label{sec:encoding}

To extract features from blurry and sharp frames, we propose an encoder architecture leveraging attention mechanisms, inspired by the \dtwonet model \citep{shang2021bringing}. However, unlike \dtwonet, which focuses on global features, we adopt a local attention strategy to better handle the non-uniform, localized blurs common in video frames. This approach not only achieves competitive results but also provides a more lightweight architecture with a faster inference time compared to \dtwonet, as demonstrated by the experiments reported in \Cref{ss:inference_time}.
To this end, the proposed encoder, detailed in the following,  integrates self-attention and cross-attention mechanisms, utilizing ResBlock \citep{shafiq2022deep}.

\begin{figure}[!h]
\centering
\includegraphics[width=0.7\columnwidth,clip, trim=0 0 0 0.2cm]{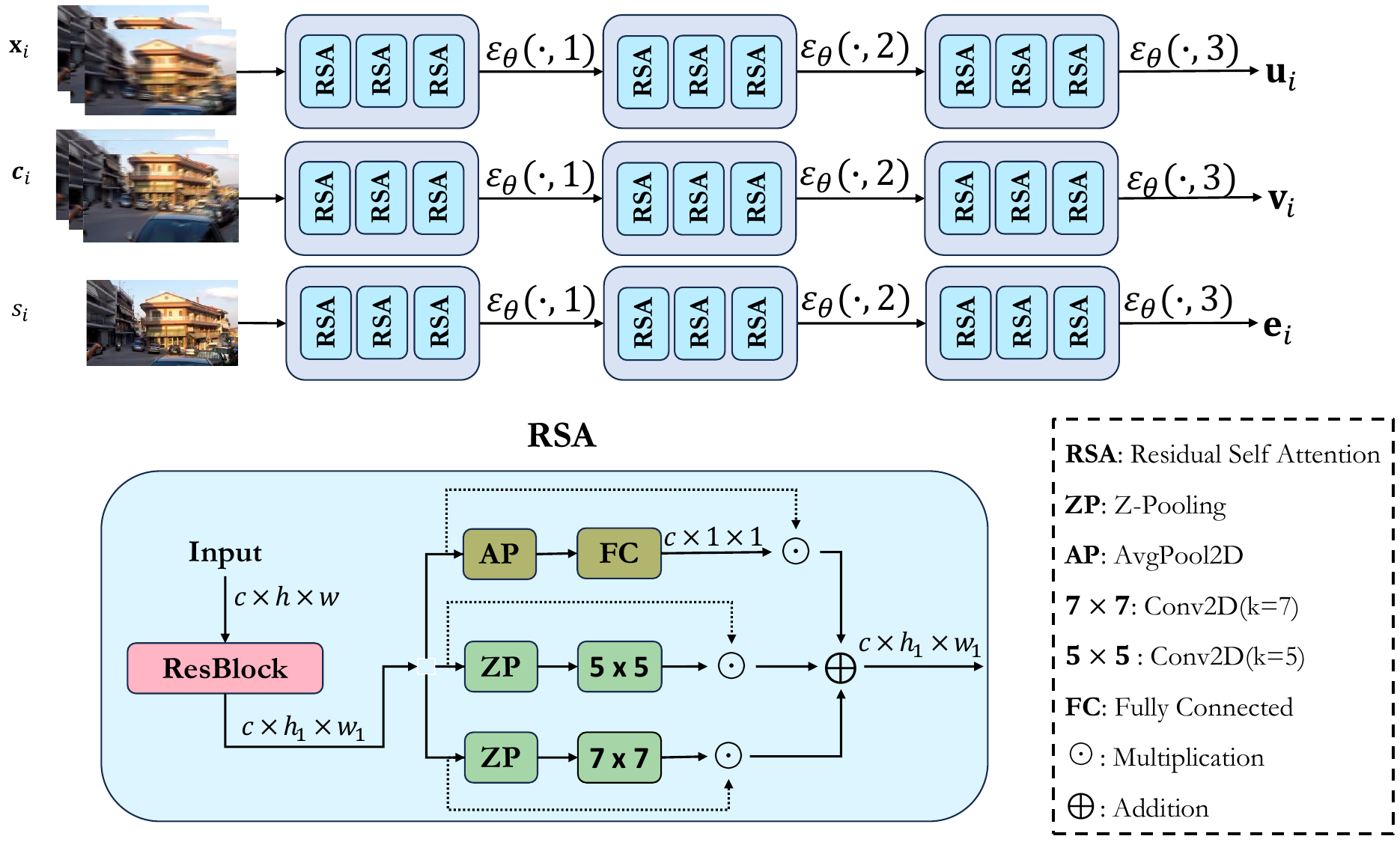}
\caption{\textit{Three blocks encoder module}. The encoder receives the blur frames \( x_i \), the emphasized edges frames \( c_i \), and any closest sharp frame $s_i$. Here \( \varepsilon_{\theta} ( \cdot, k) \) denotes the encoder until block $k$. The triplets $\uu_i$ and $\vv_i$ are deduced by applying $\varepsilon(\cdot,3)$, frame-wise, to $\xx_i$ and $\cc_i$ respectively. While, if $s_i$ is found, $\ee_i=(\varepsilon_\theta(s_i,1),\varepsilon_\theta(s_i,2),\varepsilon_\theta(s_i,3))$.}
 \label{fig:encoder}
\end{figure}

As shown in \Cref{fig:encoder}, the attention-based encoder for blur frames $F_{\text{encB}}$ receives as input the triplets of blurred frames $\xx_i$, and the $\cc_i$ sequence deduced at the previous stage to produce the two encoded sequence $\uu_i, \vv_i$. If the closest sharp frame $s_i$ is found, the $F_{\text{encS}}$ encoding module extracts the features from $s_i$ and produces the triplet $\ee_i$ in a cascade fashion, as explained in the second part of this subsection. If the sharp frames $s_i$ is not found, the $F_{\text{encS}}$ encoding module will not be executed.

The ResBlock $\varepsilon_{\theta}$ consists of three cascade-residual-blocks with attention mechanisms from shallow to deep features. We denote with $\varepsilon_{\theta}(\cdot, k)$ the network until block $k$, so that $\varepsilon_{\theta} ( \cdot, 1)$ represents the first block and $\varepsilon_{\theta} ( \cdot, 3)$ the whole encoder. Using this notation, $F_{\text{encB}}$ coincides with $\varepsilon_\theta(\cdot, 3)$ applied frame-wise to each frame of the triplet. The sharp encoder module instead is formally defined by $F_{\text{encS}}\equiv \left(\epsilon_{\theta} (\cdot, 1),\,\epsilon_{\theta} (\cdot, 2),\,\epsilon_{\theta} (\cdot, 3) \right)$. In conclusion, at this stage the following encoding are deduced
$\uu_i = \varepsilon_{\theta} (\cc_i, 3)$,
$\vv_i = \varepsilon_{\theta} (\xx_i, 3)$. Furthermore, if the sharp frame $s_i$ is found, then the following encoding is obtained $\ee_i = \left(\epsilon_{\theta} (s_i, 1),\,\epsilon_{\theta} (s_i, 2),\,\epsilon_{\theta} (s_i, 3) \right)$.

\subsubsection{Search and transfer}
\label{sec:self-search}


Depending whether sharp frames are detected, a conditional module is necessary to effectively combine information from $s_i$, when available, or process only the encoded blurred frames otherwise, as described in Equation \eqref{eq:search_and_transfer}. It is worth noting that while \dtwonet also addresses the integration of features captured by detected sharp frames, it does not include a dedicated search and transfer module. This is because the dataset \gopros, and thus also the rationale of \dtwonet, assumes a fixed frequency (\%50) of sharp frames available, meaning that, potentially, they will always be detected in the sequence. While this approach achieves high performance under such conditions, it is limited in scalability to scenarios where the availability of sharp frames varies dynamically (see experimental results). 

\begin{figure}[!h]
\centering
    \includegraphics[width=\columnwidth]{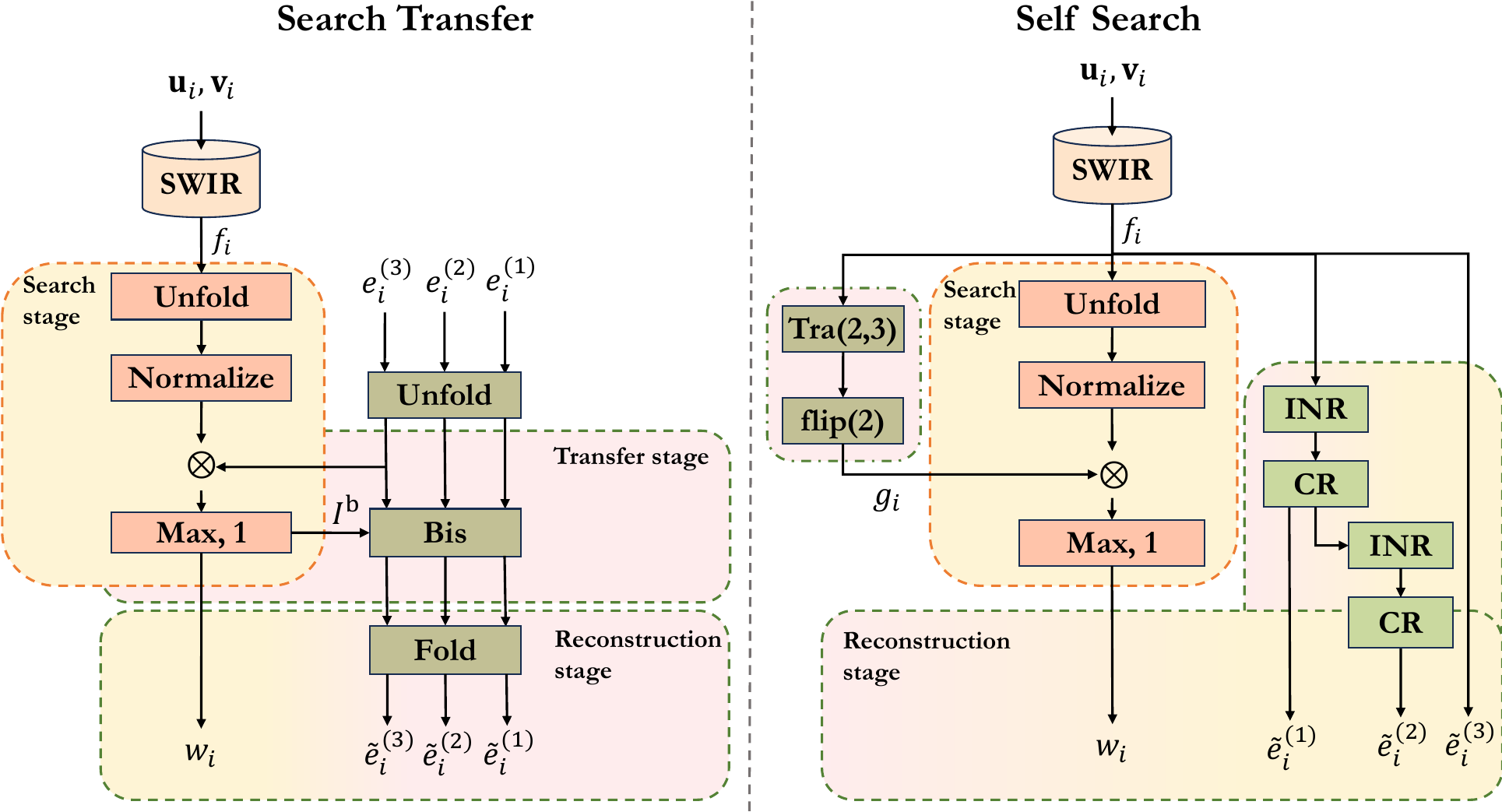}
    \caption{Detail of Search-Transfer \citep{yang2020learning} and Self-Search modules. Bis is based on the maximum similarity values and index coordinates identified in the search phase, the corresponding most similar features from the sharp features \( e_i^{(3)} \)  are selected for feature map reconstruction. The objective is to utilize the features of high-quality images to assist in the restoration or enhancement of details and textures in blurred images. Tra(2,3) denotes a transpose operation that swaps the second and third rows or columns. Flip(2) indicates a flipping operation along the second axis. INR refers to linear interpolation.}
\label{fig:search-transfer}
\end{figure}

To address this limitation, we propose a conditional Search-Transfer module. Following the scheme and operations detailed in Figure \ref{fig:search-transfer}, before either the two conditional branches (search transfer and self search), the deep features $\uu_i$ and $\vv_i$ of the blurry and sharp frame are fed into the SwinIR model \citep{liang2021swinir}. Through SwinIR, the edge and texture information of $\uu_i$ and $\vv_i$ are fused and exchanged to deduce the latent feature $f_i$ as follow
\begin{equation}
\begin{aligned}
    f^{(-)}_i &= \varphi_\theta\left((v_i + u_i), (v_{i-1} + u_{i-1})\right),\\
    f^{(+)}_i &= \varphi_\theta\left((v_i + u_i), (v_{i+1} + u_{i+1})\right),\\
    f_i &= \texttt{cat}(f^{(-)}_i, u_i, f^{(+)}_i).\\
\end{aligned}
\end{equation}
where $\varphi_\theta$ denotes the SwinIR module, \texttt{cat} the channels-wise concatenation, and $f_i$ the deep features of a blurred frame that incorporate edge information.
When the closest sharp frame \(s_i\) is not found, the Self-Search module is used, otherwise the Search-Transfer module is used.

As for the search and transfer block, we follow the operations proposed by \citep{yang2020learning} in the field of super-resolution. In this approach, deep features are first refined through a similarity analysis between $f_i$ and $ e_i$, processed in a search and transfer stages, respectively. This process produces a similarity matrix $w_i$ and a processed representation of $e_i$ that are both further refined in a final reconstruction stage.

A similar approach is applied in the self-search part; however, in this case, since no information about sharp frames is available, the transfer stage is skipped. Instead, it is replaced by a self-search mechanism, where the similarity is computed directly between features extracted from $f_i$.

\subsubsection{Attention-based decoder}

\begin{figure}[h]
\centering
\includegraphics[width=0.8\linewidth]{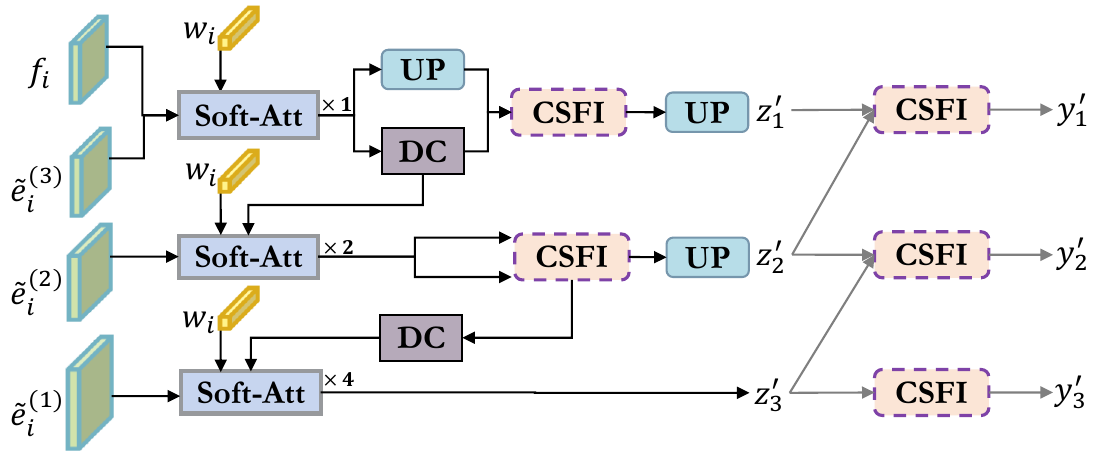}
	\caption{\small{Details of attention-based decoder. \textit{Soft-Att} refers to the Soft-Attention mechanisms \citep{yang2020learning} and \texttt{CSFI} refers to cross-scale feature integration module \citep{yang2020learning}. Combined with up-sampling modules, continually capture contextual detail information to aid in the reconstruction of blurred images.}
 }
\label{fig:decoder}
\end{figure}

The decoder part, illustrated in Figure \ref{fig:decoder}, is designed to effectively leverage the key pixel features of the sharp frame $\tilde{e}$ to restore and reconstruct the blurred features $f$ during the decoding process. Inspired by the reconstruction methods used in super-resolution models \citep{yang2020learning}, we integrated features extracted from sharp frames previously upsampled at different scales, $\tilde{e}_1$, $\tilde{e}_2$, and $\tilde{e}_3$, to enhance the blurred frames $f$ through a cascade refinement process.

Specifically, the first fusion stage (shown at the top of Figure \ref{fig:decoder}) applies cascade refinement across different scales using sharp features. This is implemented via soft-attention mechanisms \citep{yang2020learning}. The refined features are then processed using a cross-scale feature integration module \citep{yang2020learning, sun2019high}, which mixes the outcomes of the cascade stages by extracting improved combinations, resulting in $y'_1$, $y'_2$, and $y'_3$. The final output is computed as $y = y'_1 + y'_2 + y'_3$.

The scheme highlights the use of different upsampling approaches, where ``UP" represents a simple upsampling performed via interpolation, while ``DC" refers to the Resblock-based deconvolution.
The use of such super-resolution-inspired decoder blocks, combined with the processing of sharp frames, leads to improvements in accuracy, as demonstrated by the ablation studies presented in \Cref{ss:ablation_submodule}.


\section{Experiments}\label{sec:experiments}

This section reports a set of experiments aimed at showing the performance of the \speinet model on different datasets as \goproo \citep{nah2019ntire}, \gopros \citep{shang2021bringing}, \bsd \citep{zhong2020efficient}, and our proposed dataset \goprors. We also compared \speinet with the most recent state-of-the-art models, as \cdvd \citep{pan2020cascaded},  \cdvdnl \citep{pan2023cascaded}, \dstnet \citep{pan2023deep}, \vrt \citep{liang2024vrt}, \vdtr \citep{cao2022vdtr}, \dtwonet \citep{shang2021bringing}, and \pvd \citep{son2021recurrent}. A list other methods and the reasons why they are not considered in the comparison can be found in \ref{app:not-compared}.

\subsection{Dataset description}

The \goproo dataset \citep{nah2019ntire} contains $33$ video sequences with $3214$ pairs of blurred and sharp frames with a resolution of $720\times 1280$ pixels. The training set includes 22 videos, while the remaining ones are used for testing.
The \goprors dataset, introduced in \Cref{sec:dataset}, contains the same amount of videos of the \goproo dataset with same resolution. Each frame is associated with a binary label indicating whether the frame is sharp or blur. The training split of \goprors contains videos with ratios $r=0.1,0.3,0.5$ uniformly selected.
The \bsd dataset \citep{zhong2020efficient} contains $156$ real-world video sequences of $150$ frames with a resolution of $1280 \times 720$ pixels. A subset of $136$ video sequences are kept for training, while the other $20$ are used for testing.

\subsection{Experimental settings}

For training the model, we initialized the Res-Blocks of the encoder and decoder by using the initialization method described in \citep{shang2021bringing}. The Adam optimizer was used with default parameters \(\beta_1 = 0.9\), \(\beta_2 = 0.999\), and \(\epsilon = 10^{-8}\). The HEM and the L1 loss functions have been used following the settings reported in \citep{pan2023cascaded, shang2021bringing}. The batch size was set to $20$ and each token was obtained by non-overlapping square patches of size $200$. The initial learning rate was set to \(1 \times 10^{-4}\) and was multiplied by $0.5$ every $150$ epochs. The training process was terminated after $500$ epochs. For the sharp prior detector, the kernel size for the six auto-focusing measurement methods was set to $11$. Our proposed model can be trained end-to-end. The training of the model required $4$ days on $3$ NVIDIA-A100-40GB using \texttt{PyTorch1.12} for compatibility with the pre-trained inner modules.

\subsection{Comparison on different dataset}
\label{sec:comparisons}

This section presents the performance of \speinet trained on the proposed \goprors dataset, comparing it with other state-of-the-art models on \goproo, \gopros, \goprors, and \bsd datasets. In particular, Table \ref{tab:mixed_tables} reports the results of the evaluated models across all the considered datasets, with the averages highlighted at the bottom in gray.\medskip

\begin{table*}[ht]
\caption{Performance comparison of \speinet against state-of-the-art models on  \goproo, \goprors, \gopros, and BSD datasets.  \dtwonet is trained on \gopros (S) and \goprors (RS), \speinet is trained on \goprors, and the other models are trained on \goproo since do not leverage sharp frames.}
\centering
\resizebox{\textwidth}{!}{%
\begin{tabular}{l|l|cccccccc|c} 
\toprule 
\textbf{Dataset} & Metric  & \textbf{\cdvdnl} & \textbf{\vrt} & \textbf{\vdtr} & \textbf{\pvd} & \textbf{\cdvd} & \textbf{\dstnet} & \textbf{\dtwonet(S)} & \textbf{\dtwonet(RS)} & \textbf{\speinet} \\
\midrule
\textbf{GoProO} & PSNR & 32.9042 & {34.79} & 33.199 & 34.155 & 33.260 & 28.565 & 28.565 & 29.263 & 31.527 \\
& SSIM &  0.9419 & 0.9576 & 0.9407 & {0.9679} & 0.9481 & 0.8746 & 0.8746 & 0.8864 & 0.9182 \\
\midrule
\textbf{GoProS} & PSNR & 32.611 & 33.97 & 32.886 & 31.402 & 24.576 & 26.828 & {35.554} & 32.824 & 33.861 \\
& SSIM & 0.8967 & 0.9230 & 0.8962 & 0.9187 & 0.7326 & 0.8288 & {0.9491} & 0.9110 & 0.9284 \\
\midrule
\textbf{GoProRS} & PSNR & 32.608 & 33.24 & 5.96 & 32.674 & 29.840 & 28.347 & 30.103 & 30.995 & {33.396} \\
& SSIM &  0.9352 & 0.9434 & 0.0066 & 0.9562 & 0.9331 & 0.8712 & 0.8985 & 0.9077 & {0.9396} \\
\midrule
\midrule
\textbf{1ms-8ms} & PSNR & 26.573 & 24.800 & 25.050 & 23.693 & 26.375 & 25.290 & {29.750} & 29.630 & 29.601 \\
 & SSIM & 0.8148 & 0.7812 & 0.8121 & 0.7659 & 0.8228 & 0.8473 & {0.8997} & 0.8834 & 0.8883 \\
 \midrule
\textbf{2ms-16ms} & PSNR & 25.401 & 22.900 & 23.870 & 22.446 & 25.627 & 22.756 & {27.863} & 27.828 & 27.410 \\
& SSIM & 0.7887 & 0.7272 & 0.7777 & 0.6505 & 0.8016 & 0.7973 & {0.8656} & 0.8626 & 0.8498 \\
\midrule
\textbf{3ms-24ms} & PSNR & 26.522 & 24.500 & 24.920 & 23.769 & 26.030 & 23.747 & 27.845 & {28.000} & 27.547 \\
& SSIM & 0.8356 & 0.7812 & 0.8056 & 0.7807 & 0.8352 & 0.7994 & {0.8705} & 0.8739 & 0.8637 \\
\midrule
\midrule
\rowcolor[gray]{0.9}
\textbf{Average} & PSNR & 29.6033 & 28.5287 & 23.9873 & 27.7203 & 27.0622 & 26.2640 & 29.6038 & 29.7566  & \textbf{30.5570} \\
\rowcolor[gray]{0.9}
& SSIM & 0.8685  & 0.8539 & 0.7232 & 0.8797 & 0.8190 & 0.8018  & 0.8930 & 0.8875 & \textbf{0.8980} \\
\bottomrule
\end{tabular}%
}
\label{tab:mixed_tables}
\end{table*}


Regarding \dtwonet, which, like \speinet, leverages the detection of sharp frames, we compare its results when trained on our proposed dataset (\goprors) against its original version trained on \gopros. As shown in the table, the performance of the version trained on \goprors improves across nearly all datasets, indicating that the use of multiple sharp frame ratios enhances the generalization capability of the \dtwonet model. The only scenario where the original \dtwonet performs better is when evaluated on \gopros, the dataset it was originally trained on. This suggests a potential risk of overfitting in the original configuration with the \gopros dataset.

The \vrt model trained on \goproo performs better than all the ther models on this dataset. However, along with all the other models trained on the \goproo, its performance is not good when generalized to other datasets, especially on \bsd, suggesting overfitting on gopro-datasets.

The proposed \speinet model, instead, achieves the best results on the average, surpassing \dtwonet (trained on \goprors) in both PSNR ($+0.8$) and SSIM ($+0.01$). Specifically, the proposed architecture, combined with the \goprors dataset, achieves a performance that is either superior or comparable to the best results obtained by other models across all addressed datasets.

\begin{table*}[ht]
\caption{Performance comparison across \goprors dataset with different ratios. \speinet was trained on \goprors($r=0.1,\,0.3,\,0.5$) and the other method were trained on \goproo. The dag {\dag} indicates that sharp frames detection has been deduced with our detector.}
\centering
\resizebox{\textwidth}{!}{%
\begin{tabular}{lcccccccccc} 
\toprule 
\textbf{Model} & \multicolumn{2}{c}{\textbf{r=0.02}} & \multicolumn{2}{c}{\textbf{r=0.1}} & \multicolumn{2}{c}{\textbf{r=0.3}} & \multicolumn{2}{c}{\textbf{r=0.4}} & \multicolumn{2}{c}{\textbf{r=0.5}} \\ 
\cmidrule(lr){2-3} \cmidrule(lr){4-5} \cmidrule(lr){6-7} \cmidrule(lr){8-9} \cmidrule(lr){10-11}
 & \textbf{PSNR} & \textbf{SSIM} & \textbf{PSNR} & \textbf{SSIM} & \textbf{PSNR} & \textbf{SSIM} & \textbf{PSNR} & \textbf{SSIM} & \textbf{PSNR} & \textbf{SSIM} \\ 
\midrule 
\cdvdnl & 31.676 & 0.9262 & 32.212 & 0.9321 & 33.117 & 0.9377 & 33.632 & 0.9432 & 33.989 & 0.9452\\ 
\dstnet &26.851 &0.8334 & 30.515 &0.9301 & 32.224 & {0.9457} & 33.417 & {0.9601} &30.391  & 0.9184\\ 
\vrt & {33.990} & {0.9433} & {33.120} & {0.9402} & 34.020 & 0.9452 &34.380  & 0.9494 &34.750  &0.9503 \\ 
\vdtr & 5.950 & 0.0071 & 5.580 & 0.0060 & 6.280 & 0.0081 & 6.640 & 0.0105 & 5.690 &0.0064 \\ 
\cdvd & 28.122 & 0.8660 & 28.493 & 0.8737 & 28.667 & 0.8772 &28.939  & 0.8815 & 28.989 & 0.8837\\
\pvd &28.288 &0.8533 &29.091  &0.8956  &30.135  &0.9082  &30.679  &0.9166  &31.083  &0.9215 \\
\dtwonet(S) & 28.201 & 0.8581 & 29.278 &0.8814  &30.892  &0.9074  &31.551  & 0.9169 &31.939  &0.9201 \\ 
\dtwonet{$^\dagger$}(S)&28.722 &0.87  & 30.384& 0.9049&31.761 &0.9213 &32.291&0.9287&32.691&0.9319\\ 
\dtwonet(RS) & 29.659 & 0.8906 & 30.307 & 0.9000 & 31.882 & 0.9154 &32.563 & 0.9238 & 33.010  & 0.9278 \\ 
\midrule
\textbf{\speinet} &32.032 &0.9277  &32.625  &0.9334  &{34.295}  & 0.9448 &{35.029} & 0.9509 & {35.452}  & {0.9536} \\ 
\bottomrule 
\end{tabular}}

\label{tab:comparison3}
\end{table*}

\subsection{Comparisons with different sharp-ratios}

\Cref{tab:comparison3} provides a detailed analysis of \speinet and other models on the \goprors test sets, considering different sharp frame ratios. Notably, our method achieves the highest PSNR as the number of sharp frames increases (above $r = 0.3$). Contrary to expectations, the other models do not exhibit significant improvements in restoration performance on video sequences with a high number of sharp frames.
The \dtwonet model trained on \gopros, designed to exploit sharp frame information, shows a slight improvement in PSNR as the number of sharp frames increases. However, this improvement follows a less pronounced trend compared to \speinet. Additionally, \dtwonet benefits significantly from higher values of $r$ when trained on our proposed \goprors dataset, highlighting the advantages of effectively addressing a wider range of sharp frame ratios.

\section{Ablation study} \label{sec:ablation}

The aim of this section is to prove that the sub-modules proposed in the design of the \speinet are helpful for improving the performance, while also reducing the inference cost with respect to related work.

\subsection{Sharp frames detection}
\label{sec:abl-detection}

This section aims at better understanding the performance of the proposed detection method when applied to deblurring models that leverage sharp frames. Specifically, we focus on \dtwonet and \speinet, as both utilize the detection of sharp frames, unlike other models discussed in the previous section. The quality of the proposed detector is assessed by comparing it with the \texttt{Bi-LSTM} method (presented and used in \dtwonet). The \texttt{Bi-LSTM} is evaluated in its original version trained on the \gopros dataset, as the authors did not provide the training files for that model. In contrast, the proposed detector is trained both on \goprors and \gopros.

\Cref{tab:albations1} presents evaluations conducted on \goprors with $r=0.5$, where the logistic-regression-based detector (LD) is also trained. As shown in the table, LD outperforms \texttt{Bi-LSTM}, improving detection accuracy by 15.56\%, which is partiality also due to the different training set used for \texttt{Bi-LSTM}. 
In terms of inference time, LD processes a \goprors video in an average of $5.71$ seconds, compared to \texttt{Bi-LSTM}'s $46$ seconds. Using \texttt{Bi-LSTM} would increase the total inference time of \speinet by $13.56\%$ for processing a single video sequence.
This is because LD has a more lightweight designed as it is based on logistic regression applied to six metrics (\Cref{sec:sharp-frame-detector}), rather than a large LSTM model. 

\begin{table}[ht]
\caption{Performance of the \speinet adopting LD and Bi-LSTM detectors compared on the \goprors(r=0.5) dataset. The Bi-LSTM was trained on \gopros, whereas LD was trained on \goprors.}
\centering
\begin{tabular}{lccc} 
\toprule 
Method  & Acc.[\%] & Time [s]
& PSNR/SSIM \\
\midrule
LD &75.95 & 5.71
& 35.452/0.9536 \\ 
Bi-LSTM & 60.39 & 46 (+13.56\%) &  35.423/0.9535 \\ 
\bottomrule
\end{tabular}
\label{tab:albations1}
\end{table}
Furthermore, considering a more fair comparison,  \Cref{tab:albationsgopros} reports the evaluation based on \gopros. In this case, \texttt{Bi-LSTM} achieves better accuracy due to its high performance on this scenario. However, its significantly larger inference time justifies the use of LD to achieve a better trade-off between performance and computational cost. In fact, although the detection accuracy is lower than 2.56\%, in terms of prediction time, LD takes an average of $6.31$ s to process a \gopros video in average, while \texttt{Bi-LSTM} takes $50.83$ s. If \texttt{Bi-LSTM} is used instead, \dtwonet's entire inference will take $12.51\%$ to process a single video sequence.




\begin{table}[ht]
\caption{\small{Performance of the \dtwonet adopting LD and Bi-LSTM detectors compared on the \gopros dataset. Both the \texttt{Bi-LST} and LD were trained on \gopros.}}
\centering
\begin{tabular}{lccc} 
\toprule 
Method  & Acc.[\%] & Time [s]
& PSNR/SSIM \\
\midrule
LD &95.85 & 6.31
&35.552/0.9485\\ 
Bi-LSTM & 98.41 & 50.83 (+12.51\%) &  35.554/0.9491  \\ 
\bottomrule
\end{tabular}
\label{tab:albationsgopros}
\end{table}

To summarize, we acknowledge that a slight drop in the detection accuracy of sharp frames results in minimal variation in PSNR and SSIM, suggesting that LD is a viable option when computational efficiency is a priority. 


\subsection{\speinet sub-modules}
\label{ss:ablation_submodule}


We conduct ablation studies to investigate the importance of the various submodules of \speinet, and the results are reported in \Cref{tab:ablation-submodules}. In detail, we evaluated the performance of the proposed model by gradually activating the $F_S$ and $F_{REE}$ modules, and the multi-scaling augmentation of the decoder stage. The deactivation of the $F_S$ module can easily be done by forcing the output $t_i$ of the detector to be $-1$ for each triplet. The deactivation of the $F_{REE}$ module can be done by avoiding merging the information of $v_i$, thus creating $f_i$ only from $u_i$ through channel-wise concatenation. 

\begin{table}[!ht]
\caption{Performance of \speinet for different sub-module configurations, all trained on the \goprors dataset sampling uniformly three different ratios: 0.1, 0.3, and 0.5. The metrics are computed on the test set of $\goprors(0.3)$. \textbf{\(F_{\text{dec}}\)} refers to the decoder with the multi-scale augmentation strategy.}
\centering

\begin{tabular}{ccccc} 
\toprule 
\textbf{Sharp}  &\textbf{REE} 
&\textbf{\(F_{\text{dec}}\)} & PSNR&SSIM\\
\midrule
\ding{55} &\ding{55}  &\ding{55} & 29.592 & 0.8620  \\ 
\ding{55} &\ding{55}  &\checkmark & 29.684 (+0.092\%) &0.8640 \\ 
\ding{55} &\checkmark  &\checkmark & 33.953 (+4.361\%)&0.9359 \\ 
\checkmark & \ding{55}  &\checkmark & 33.854 (+4.262\%)&0.9422 \\ 
\midrule
\checkmark & \checkmark & \checkmark & \textbf{34.295 (+4.703\%)}& 0.9448\\ 
\bottomrule 
\end{tabular}
\label{tab:ablation-submodules}
\end{table}

As can be seen from the table, the prior detection of sharp frames greatly improves the video deblurring performance, which is 4.262\% higher than the baseline attention encoder-decoder structure. The edge information of REE greatly helps the edge texture reconstruction work during decoding and reconstruction, which is 4.361\% higher. Therefore, the construction of sharp prior detection and edge information modules greatly improves the performance of video deblurring. The reconstruction of key detail pixel information and edge texture of sharp frames is crucial for recovering and reconstructing blurred frames.

\subsection{Inference time of video deblurring models}
\label{ss:inference_time}
Considering the \goprors dataset, we compared the computational time of each test model in \Cref{tab:timing}. As shown in the table, when considering the most performant models reported in \Cref{tab:mixed_tables}, \speinet demonstrates not only superior performance but also greater efficiency compared to other competitors. For instance, as discussed in Section~\ref{sec:speinet}, the specific modules adopted in \speinet, as opposed to those used in \dtwonet, make \speinet both more lightweight and more generalizable, in conjunction with the adoption of the proposed dataset, to varying availability of sharp frames.

\begin{table*}[!htbp]
\caption{Average Inference time in minutes to process one video-sequence.}
\centering
\resizebox{\textwidth}{!}{%
\begin{tabular}{lccccccccc} 
\toprule
\textbf{Dataset} &\textbf{\cdvdnl}  & \textbf{\dtwonet(S)} & \textbf{\vrt} & \textbf{\vdtr} & \textbf{\dstnet} &\textbf{\cdvd} &\textbf{\pvd}&\textbf{\speinet}\\ 
\midrule
\textbf{\goprors(0.2)}  &3.0  &7.24  &32.4  &3.0  &1.2  &7.2 &0.24 & 4.95  \\

\bottomrule 
\end{tabular}}
\label{tab:timing}

\end{table*}
\section{Visualization and discussion}
\label{sec:visualization}

\begin{figure*}[!h]
    \centering
    \includegraphics[width=\linewidth]{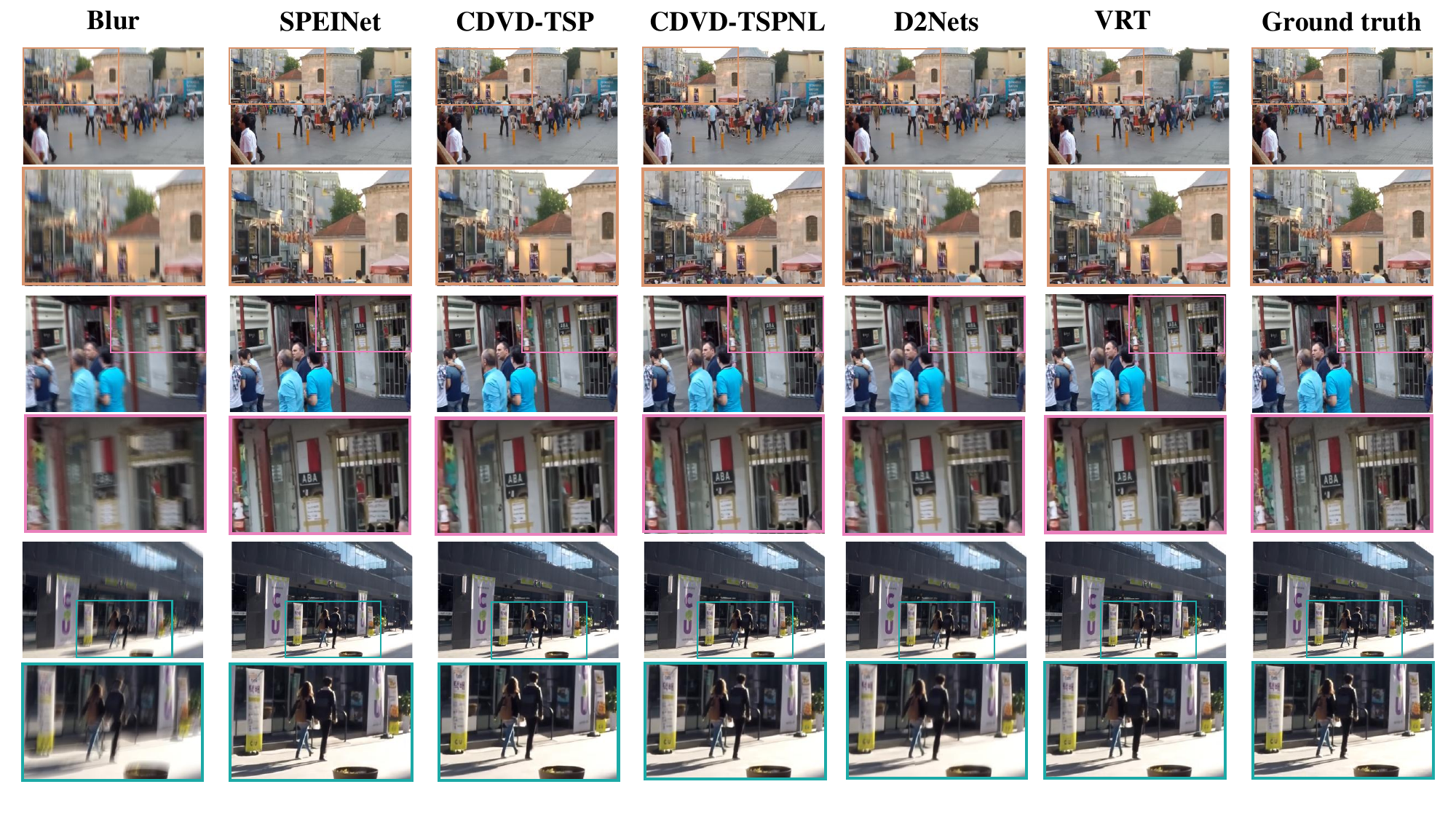}
    \caption{Sample outputs produced by \speinet and other state-of-the-art models on six frames taken from \goprors(r=0.02).}
    \label{fig:visualization1}
\end{figure*}

This section aims at visually inspecting the quality of results produced by the \speinet model compared with the other models. 
The \speinet exhibits a great capability of recovering the details of the blurred frames, even when the sharp frames are only 2\% of the whole video. \Cref{fig:visualization1} illustrates a few examples of outputs produced by \speinet and other state-of-the-art models on six blur frames taken from \goprors setting $r$ = 0.02.

\begin{figure*}[!h]
    \centering
    \includegraphics[width=\linewidth]{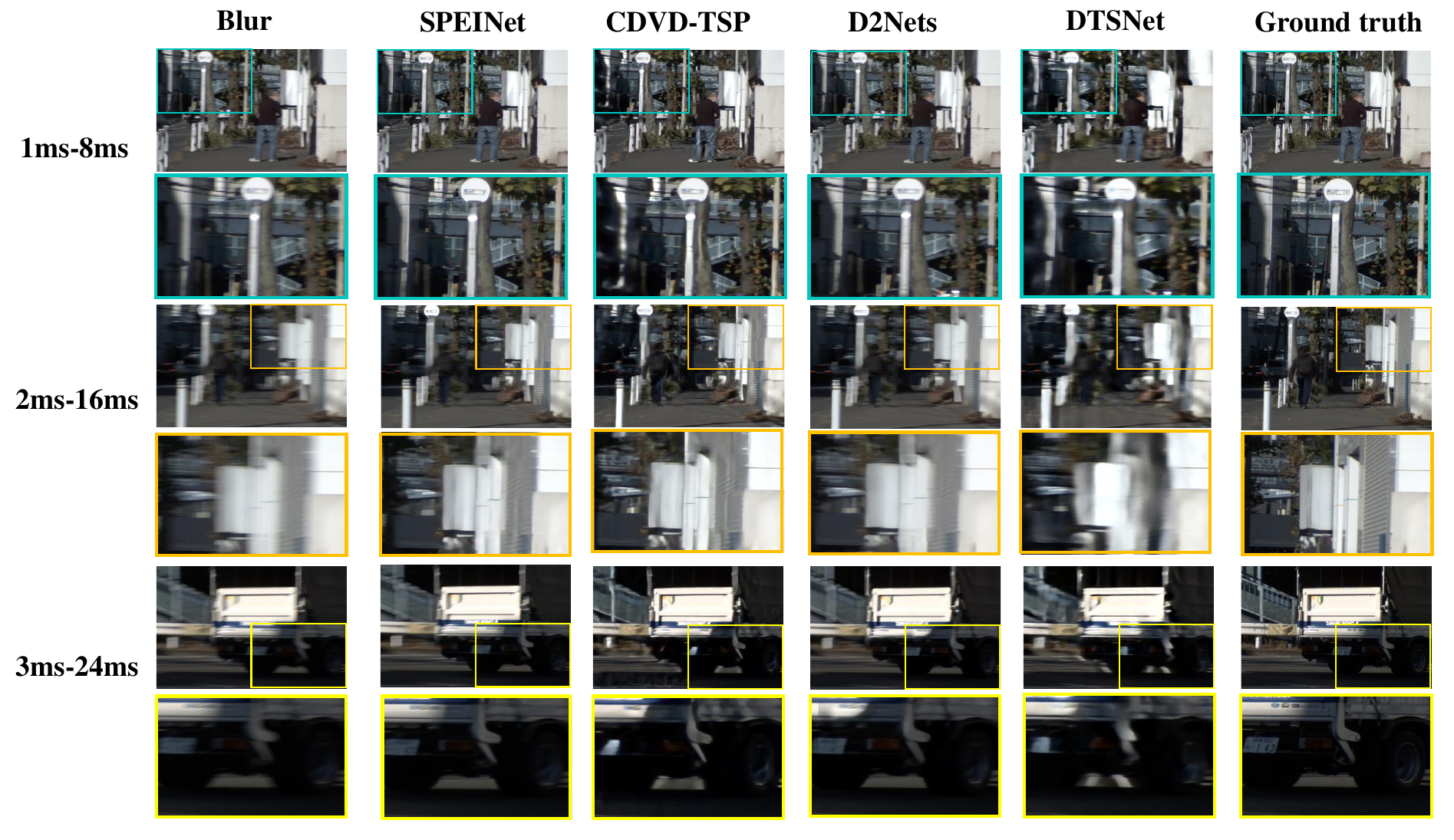}
    \caption{Examples of outputs produced by \speinet and other state-of-the-art models on three samples sequences taken from the \bsd real-world dataset.}
    \label{fig:visualization2}
\end{figure*}

Compared to \cdvd, \speinet is capable of reconstructing finer details, such as the text in the wall shown in the second row of the figure.
Comparing to \vrt, the \speinet model better recovers the facial contours of the crowd, especially the building  details, as visible in the first row.
Compared to \cdvdnl, the \speinet model has a greater capacity in recovering the motion blur of the people's hair very well, as shown in the third row.

Furthermore, \Cref{fig:visualization2} shows a comparison of the deblurring capabilities of the models on $3$ samples of the \bsd real-world dataset. As shown in the image, even in case of non-uniform motion blur, \speinet is capable of recovering edge details of the images, which has strong real-world application capabilities.

\section{Conclusions} \label{sec:conclusion}

This work presented a novel video deblurring model called Sharpness Prior Detector and Edge Information Network (\speinet), which exploits the information retained in the sharp frames to improve the quality of deblurring. In order to avoid additional computational overhead, the detection of sharp frames uses an logistic-regression-based detector (LD), which has been shown to have a better performance with respect to a \texttt{Bi-LSTM}. To achieve a better model performance during training, we also proposed the \goprors dataset, containing a user-defined ratio of sharp/blur frames.\medskip

Extensive experiments have been conducted on \goproo, \goprors, \gopros, and \bsd as real-world reference, comparing the performance of \speinet against a number of state-of-the-art methods. The results show that the proposed \speinet not only achieves remarkable video deblurring results, but also has strong domain adaptability and can effectively be applied to real-world deblurring scenarios. \speinet achieves an average PSNR of $30.557$ on the tested datasets, which represents an improvement of over $2.74\%$ compared to the PSNR achieved by other models.\medskip

As a future work, we plan to improve the sharp detection stage of \speinet,
since it is fundamental for a better performance of the model. In the present model, the fitting of the LD detector is done in a supervised fashion using the ground truth labels available in the dataset. In a future work, we will consider the possibility of using the information contained in the sharp frames, leaving the model to judge which information can be retained for improving the deblurring result.

\subsubsection*{Data Availability}
Code and data will be publicly available at \href{{https://github.com/yangt1013/SPEINet}}{\tt github.com/yangt1013/SPEINet}

\bibliographystyle{elsarticle-harv} 
\bibliography{bibliography}

\begin{thebibliography}{56}
\expandafter\ifx\csname natexlab\endcsname\relax\def\natexlab#1{#1}\fi
\providecommand{\url}[1]{\texttt{#1}}
\providecommand{\href}[2]{#2}
\providecommand{\path}[1]{#1}
\providecommand{\DOIprefix}{doi:}
\providecommand{\ArXivprefix}{arXiv:}
\providecommand{\URLprefix}{URL: }
\providecommand{\Pubmedprefix}{pmid:}
\providecommand{\doi}[1]{\href{http://dx.doi.org/#1}{\path{#1}}}
\providecommand{\Pubmed}[1]{\href{pmid:#1}{\path{#1}}}
\providecommand{\bibinfo}[2]{#2}
\ifx\xfnm\relax \def\xfnm[#1]{\unskip,\space#1}\fi
\bibitem[{Amini~Gougeh et~al.(2021)Amini~Gougeh, Yousefi~Rezaii and Farzamnia}]{amini2021medical}
\bibinfo{author}{Amini~Gougeh, R.}, \bibinfo{author}{Yousefi~Rezaii, T.}, \bibinfo{author}{Farzamnia, A.}, \bibinfo{year}{2021}.
\newblock \bibinfo{title}{Medical image enhancement and deblurring}, in: \bibinfo{booktitle}{Proceedings of the 11th National Technical Seminar on Unmanned System Technology 2019: NUSYS'19}, \bibinfo{publisher}{Springer Singapore}. pp. \bibinfo{pages}{543--554}.
\bibitem[{Cao et~al.(2022)Cao, Fan, Zhang, Wang and Yang}]{cao2022vdtr}
\bibinfo{author}{Cao, M.}, \bibinfo{author}{Fan, Y.}, \bibinfo{author}{Zhang, Y.}, \bibinfo{author}{Wang, J.}, \bibinfo{author}{Yang, Y.}, \bibinfo{year}{2022}.
\newblock \bibinfo{title}{Vdtr: Video deblurring with transformer}.
\newblock \bibinfo{journal}{IEEE Transactions on Circuits and Systems for Video Technology} \bibinfo{volume}{33}, \bibinfo{pages}{160--171}.
\bibitem[{Chakravarthi et~al.(2024)Chakravarthi, Verma, Daniilidis, Fermuller and Yang}]{chakravarthi2024recent}
\bibinfo{author}{Chakravarthi, B.}, \bibinfo{author}{Verma, A.A.}, \bibinfo{author}{Daniilidis, K.}, \bibinfo{author}{Fermuller, C.}, \bibinfo{author}{Yang, Y.}, \bibinfo{year}{2024}.
\newblock \bibinfo{title}{Recent event camera innovations: A survey}.
\newblock \bibinfo{journal}{arXiv preprint arXiv:2408.13627} .
\bibitem[{Chen et~al.(2024)Chen, Zhang, Liu, Gu, Kong and Yuan}]{chen2024hierarchical}
\bibinfo{author}{Chen, Z.}, \bibinfo{author}{Zhang, Y.}, \bibinfo{author}{Liu, D.}, \bibinfo{author}{Gu, J.}, \bibinfo{author}{Kong, L.}, \bibinfo{author}{Yuan, X.}, \bibinfo{year}{2024}.
\newblock \bibinfo{title}{Hierarchical integration diffusion model for realistic image deblurring}, in: \bibinfo{booktitle}{Advances in Neural Information Processing Systems}.
\bibitem[{Cheng et~al.(2022)Cheng, Zhu, Zhan and Pei}]{cheng2022video}
\bibinfo{author}{Cheng, K.}, \bibinfo{author}{Zhu, X.}, \bibinfo{author}{Zhan, Y.}, \bibinfo{author}{Pei, Y.}, \bibinfo{year}{2022}.
\newblock \bibinfo{title}{Video deblurring and flow-guided feature aggregation for obstacle detection in agricultural videos}.
\newblock \bibinfo{journal}{International Journal of Multimedia Information Retrieval} \bibinfo{volume}{11}, \bibinfo{pages}{577--588}.
\bibitem[{Cui et~al.(2023)Cui, Tao, Ren and Knoll}]{cui2023dual}
\bibinfo{author}{Cui, Y.}, \bibinfo{author}{Tao, Y.}, \bibinfo{author}{Ren, W.}, \bibinfo{author}{Knoll, A.}, \bibinfo{year}{2023}.
\newblock \bibinfo{title}{Dual-domain attention for image deblurring}, in: \bibinfo{booktitle}{Proceedings of the AAAI Conference on Artificial Intelligence}, pp. \bibinfo{pages}{479--487}.
\bibitem[{Delbracio and Sapiro(2015)}]{delbracio2015handheld}
\bibinfo{author}{Delbracio, M.}, \bibinfo{author}{Sapiro, G.}, \bibinfo{year}{2015}.
\newblock \bibinfo{title}{Hand-held video deblurring via efficient fourier aggregation}.
\newblock \bibinfo{journal}{IEEE Transactions on Computational Imaging} \bibinfo{volume}{1}, \bibinfo{pages}{270--283}.
\bibitem[{Dong et~al.(2023)Dong, Pan, Yang and Tang}]{dong2023multi}
\bibinfo{author}{Dong, J.}, \bibinfo{author}{Pan, J.}, \bibinfo{author}{Yang, Z.}, \bibinfo{author}{Tang, J.}, \bibinfo{year}{2023}.
\newblock \bibinfo{title}{Multi-scale residual low-pass filter network for image deblurring}, in: \bibinfo{booktitle}{Proceedings of the IEEE/CVF International Conference on Computer Vision}, pp. \bibinfo{pages}{12345--12354}.
\bibitem[{He et~al.(2025)He, Tsai, Wu, Peng, Tsai, Lin and Lin}]{he2025domain}
\bibinfo{author}{He, J.T.}, \bibinfo{author}{Tsai, F.J.}, \bibinfo{author}{Wu, J.H.}, \bibinfo{author}{Peng, Y.T.}, \bibinfo{author}{Tsai, C.C.}, \bibinfo{author}{Lin, C.W.}, \bibinfo{author}{Lin, Y.Y.}, \bibinfo{year}{2025}.
\newblock \bibinfo{title}{Domain-adaptive video deblurring via test-time blurring}, in: \bibinfo{booktitle}{European Conference on Computer Vision}, \bibinfo{organization}{Springer}. pp. \bibinfo{pages}{125--142}.
\bibitem[{Huang et~al.(2024)Huang, Chen, Chen, Li, Hsiao, Wang and Wen}]{huang2024effective}
\bibinfo{author}{Huang, D.Y.}, \bibinfo{author}{Chen, C.H.}, \bibinfo{author}{Chen, T.Y.}, \bibinfo{author}{Li, J.E.}, \bibinfo{author}{Hsiao, H.L.}, \bibinfo{author}{Wang, D.J.}, \bibinfo{author}{Wen, C.K.}, \bibinfo{year}{2024}.
\newblock \bibinfo{title}{Effective video deblurring based on feature-enhanced deep learning network for daytime and nighttime images}.
\newblock \bibinfo{journal}{Multimedia Tools and Applications} , \bibinfo{pages}{1--27}.
\bibitem[{Hyun~Kim and Mu~Lee(2015)}]{kim2015generalized}
\bibinfo{author}{Hyun~Kim, T.}, \bibinfo{author}{Mu~Lee, K.}, \bibinfo{year}{2015}.
\newblock \bibinfo{title}{Generalized video deblurring for dynamic scenes}, in: \bibinfo{booktitle}{Proceedings of the IEEE Conference on Computer Vision and Pattern Recognition}, pp. \bibinfo{pages}{5426--5434}.
\bibitem[{Imani et~al.(2024)Imani, Islam, Junayed and Ahad}]{imani2024stereoscopic}
\bibinfo{author}{Imani, H.}, \bibinfo{author}{Islam, M.B.}, \bibinfo{author}{Junayed, M.S.}, \bibinfo{author}{Ahad, M.A.R.}, \bibinfo{year}{2024}.
\newblock \bibinfo{title}{Stereoscopic video deblurring transformer}.
\newblock \bibinfo{journal}{Scientific Reports} \bibinfo{volume}{14}, \bibinfo{pages}{14342}.
\bibitem[{Kim et~al.(2024)Kim, Cho and Yoon}]{kim2024frequency}
\bibinfo{author}{Kim, T.}, \bibinfo{author}{Cho, H.}, \bibinfo{author}{Yoon, K.J.}, \bibinfo{year}{2024}.
\newblock \bibinfo{title}{Frequency-aware event-based video deblurring for real-world motion blur}, in: \bibinfo{booktitle}{Proceedings of the IEEE/CVF Conference on Computer Vision and Pattern Recognition}, pp. \bibinfo{pages}{24966--24976}.
\bibitem[{Kim et~al.(2022)Kim, Lee, Wang and Yoon}]{kim2022event}
\bibinfo{author}{Kim, T.}, \bibinfo{author}{Lee, J.}, \bibinfo{author}{Wang, L.}, \bibinfo{author}{Yoon, K.J.}, \bibinfo{year}{2022}.
\newblock \bibinfo{title}{Event-guided deblurring of unknown exposure time videos}, in: \bibinfo{booktitle}{European Conference on Computer Vision}, \bibinfo{publisher}{Springer Nature Switzerland}, \bibinfo{address}{Cham}. pp. \bibinfo{pages}{519--538}.
\bibitem[{Kong et~al.(2023)Kong, Dong, Ge, Li and Pan}]{kong2023efficient}
\bibinfo{author}{Kong, L.}, \bibinfo{author}{Dong, J.}, \bibinfo{author}{Ge, J.}, \bibinfo{author}{Li, M.}, \bibinfo{author}{Pan, J.}, \bibinfo{year}{2023}.
\newblock \bibinfo{title}{Efficient frequency domain-based transformers for high-quality image deblurring}, in: \bibinfo{booktitle}{Proceedings of the IEEE/CVF Conference on Computer Vision and Pattern Recognition}, pp. \bibinfo{pages}{5886--5895}.
\bibitem[{Liang et~al.(2024)Liang, Cao, Fan, Zhang, Ranjan, Li, Timofte and Van~Gool}]{liang2024vrt}
\bibinfo{author}{Liang, J.}, \bibinfo{author}{Cao, J.}, \bibinfo{author}{Fan, Y.}, \bibinfo{author}{Zhang, K.}, \bibinfo{author}{Ranjan, R.}, \bibinfo{author}{Li, Y.}, \bibinfo{author}{Timofte, R.}, \bibinfo{author}{Van~Gool, L.}, \bibinfo{year}{2024}.
\newblock \bibinfo{title}{Vrt: A video restoration transformer}.
\newblock \bibinfo{journal}{IEEE Transactions on Image Processing} .
\bibitem[{Liang et~al.(2021)Liang, Cao, Sun, Zhang, Van~Gool and Timofte}]{liang2021swinir}
\bibinfo{author}{Liang, J.}, \bibinfo{author}{Cao, J.}, \bibinfo{author}{Sun, G.}, \bibinfo{author}{Zhang, K.}, \bibinfo{author}{Van~Gool, L.}, \bibinfo{author}{Timofte, R.}, \bibinfo{year}{2021}.
\newblock \bibinfo{title}{Swinir: Image restoration using swin transformer}, in: \bibinfo{booktitle}{Proceedings of the IEEE/CVF International Conference on Computer Vision}, pp. \bibinfo{pages}{1833--1844}.
\bibitem[{Lin et~al.(2024)Lin, Wei, Liu, Feng and Zhao}]{lin2024lightvid}
\bibinfo{author}{Lin, L.}, \bibinfo{author}{Wei, G.}, \bibinfo{author}{Liu, K.}, \bibinfo{author}{Feng, W.}, \bibinfo{author}{Zhao, T.}, \bibinfo{year}{2024}.
\newblock \bibinfo{title}{Lightvid: Efficient video deblurring with spatial-temporal feature fusion}.
\newblock \bibinfo{journal}{IEEE Transactions on Circuits and Systems for Video Technology} .
\bibitem[{Lin et~al.(2020)Lin, Zhang, Pan, Jiang, Zou, Wang and Ren}]{lin2020learning}
\bibinfo{author}{Lin, S.}, \bibinfo{author}{Zhang, J.}, \bibinfo{author}{Pan, J.}, \bibinfo{author}{Jiang, Z.}, \bibinfo{author}{Zou, D.}, \bibinfo{author}{Wang, Y.}, \bibinfo{author}{Ren, J.}, \bibinfo{year}{2020}.
\newblock \bibinfo{title}{Learning event-driven video deblurring and interpolation}, in: \bibinfo{booktitle}{Computer Vision -- ECCV 2020: 16th European Conference, Glasgow, UK, August 23–28, 2020, Proceedings, Part VIII}, \bibinfo{publisher}{Springer International Publishing}. pp. \bibinfo{pages}{695--710}.
\bibitem[{Mao et~al.(2023)Mao, Liu, Liu, Li, Shen and Wang}]{mao2023intriguing}
\bibinfo{author}{Mao, X.}, \bibinfo{author}{Liu, Y.}, \bibinfo{author}{Liu, F.}, \bibinfo{author}{Li, Q.}, \bibinfo{author}{Shen, W.}, \bibinfo{author}{Wang, Y.}, \bibinfo{year}{2023}.
\newblock \bibinfo{title}{Intriguing findings of frequency selection for image deblurring}, in: \bibinfo{booktitle}{Proceedings of the AAAI Conference on Artificial Intelligence}, pp. \bibinfo{pages}{1905--1913}.
\bibitem[{Marr and Hildreth(1980)}]{marr1980theory}
\bibinfo{author}{Marr, D.}, \bibinfo{author}{Hildreth, E.}, \bibinfo{year}{1980}.
\newblock \bibinfo{title}{Theory of edge detection}.
\newblock \bibinfo{journal}{Proceedings of the Royal Society of London. Series B. Biological Sciences} \bibinfo{volume}{207}, \bibinfo{pages}{187--217}.
\bibitem[{Nah et~al.(2019)Nah, Baik, Hong, Moon, Son, Timofte and Mu~Lee}]{nah2019ntire}
\bibinfo{author}{Nah, S.}, \bibinfo{author}{Baik, S.}, \bibinfo{author}{Hong, S.}, \bibinfo{author}{Moon, G.}, \bibinfo{author}{Son, S.}, \bibinfo{author}{Timofte, R.}, \bibinfo{author}{Mu~Lee, K.}, \bibinfo{year}{2019}.
\newblock \bibinfo{title}{Ntire 2019 challenge on video deblurring and super-resolution: Dataset and study}, in: \bibinfo{booktitle}{Proceedings of the IEEE/CVF Conference on Computer Vision and Pattern Recognition Workshops}, pp. \bibinfo{pages}{0--0}.
\bibitem[{Nah et~al.(2017)Nah, Hyun~Kim and Mu~Lee}]{nah2017deep}
\bibinfo{author}{Nah, S.}, \bibinfo{author}{Hyun~Kim, T.}, \bibinfo{author}{Mu~Lee, K.}, \bibinfo{year}{2017}.
\newblock \bibinfo{title}{Deep multi-scale convolutional neural network for dynamic scene deblurring}, in: \bibinfo{booktitle}{Proceedings of the IEEE Conference on Computer Vision and Pattern Recognition}, pp. \bibinfo{pages}{3883--3891}.
\bibitem[{Nayak et~al.(2021)Nayak, Pati and Das}]{nayak2021comprehensive}
\bibinfo{author}{Nayak, R.}, \bibinfo{author}{Pati, U.C.}, \bibinfo{author}{Das, S.K.}, \bibinfo{year}{2021}.
\newblock \bibinfo{title}{A comprehensive review on deep learning-based methods for video anomaly detection}.
\newblock \bibinfo{journal}{Image and Vision Computing} \bibinfo{volume}{106}, \bibinfo{pages}{104078}.
\bibitem[{Pan et~al.(2020)Pan, Bai and Tang}]{pan2020cascaded}
\bibinfo{author}{Pan, J.}, \bibinfo{author}{Bai, H.}, \bibinfo{author}{Tang, J.}, \bibinfo{year}{2020}.
\newblock \bibinfo{title}{Cascaded deep video deblurring using temporal sharpness prior}, in: \bibinfo{booktitle}{Proceedings of the IEEE/CVF Conference on Computer Vision and Pattern Recognition}, pp. \bibinfo{pages}{3043--3051}.
\bibitem[{Pan et~al.(2023a)Pan, Xu, Bai, Tang and Yang}]{pan2023cascaded}
\bibinfo{author}{Pan, J.}, \bibinfo{author}{Xu, B.}, \bibinfo{author}{Bai, H.}, \bibinfo{author}{Tang, J.}, \bibinfo{author}{Yang, M.H.}, \bibinfo{year}{2023}a.
\newblock \bibinfo{title}{Cascaded deep video deblurring using temporal sharpness prior and non-local spatial-temporal similarity}.
\newblock \bibinfo{journal}{IEEE Transactions on Pattern Analysis and Machine Intelligence} \bibinfo{volume}{45}, \bibinfo{pages}{9411--9425}.
\bibitem[{Pan et~al.(2023b)Pan, Xu, Dong, Ge and Tang}]{pan2023deep}
\bibinfo{author}{Pan, J.}, \bibinfo{author}{Xu, B.}, \bibinfo{author}{Dong, J.}, \bibinfo{author}{Ge, J.}, \bibinfo{author}{Tang, J.}, \bibinfo{year}{2023}b.
\newblock \bibinfo{title}{Deep discriminative spatial and temporal network for efficient video deblurring}, in: \bibinfo{booktitle}{Proceedings of the IEEE/CVF Conference on Computer Vision and Pattern Recognition}, pp. \bibinfo{pages}{22191--22200}.
\bibitem[{Pan et~al.(2017)Pan, Dai, Liu and Porikli}]{pan2017simultaneous}
\bibinfo{author}{Pan, L.}, \bibinfo{author}{Dai, Y.}, \bibinfo{author}{Liu, M.}, \bibinfo{author}{Porikli, F.}, \bibinfo{year}{2017}.
\newblock \bibinfo{title}{Simultaneous stereo video deblurring and scene flow estimation}, in: \bibinfo{booktitle}{Proceedings of the IEEE Conference on Computer Vision and Pattern Recognition}, pp. \bibinfo{pages}{4382--4391}.
\bibitem[{Pertuz et~al.(2013)Pertuz, Puig and Garcia}]{pertuz2013analysis}
\bibinfo{author}{Pertuz, S.}, \bibinfo{author}{Puig, D.}, \bibinfo{author}{Garcia, M.A.}, \bibinfo{year}{2013}.
\newblock \bibinfo{title}{Analysis of focus measure operators for shape-from-focus}.
\newblock \bibinfo{journal}{Pattern Recognition} \bibinfo{volume}{46}, \bibinfo{pages}{1415--1432}.
\bibitem[{Rao et~al.(2025)Rao, Li, Lan, Sun, Luan, Xing, Zhao, Lin, Dong and Zhang}]{rao2025rethinking}
\bibinfo{author}{Rao, C.}, \bibinfo{author}{Li, G.}, \bibinfo{author}{Lan, Z.}, \bibinfo{author}{Sun, J.}, \bibinfo{author}{Luan, J.}, \bibinfo{author}{Xing, W.}, \bibinfo{author}{Zhao, L.}, \bibinfo{author}{Lin, H.}, \bibinfo{author}{Dong, J.}, \bibinfo{author}{Zhang, D.}, \bibinfo{year}{2025}.
\newblock \bibinfo{title}{Rethinking video deblurring with wavelet-aware dynamic transformer and diffusion model}, in: \bibinfo{booktitle}{European Conference on Computer Vision}, \bibinfo{organization}{Springer}. pp. \bibinfo{pages}{421--437}.
\bibitem[{Ren et~al.(2023)Ren, Delbracio, Talebi, Gerig and Milanfar}]{ren2023multiscale}
\bibinfo{author}{Ren, M.}, \bibinfo{author}{Delbracio, M.}, \bibinfo{author}{Talebi, H.}, \bibinfo{author}{Gerig, G.}, \bibinfo{author}{Milanfar, P.}, \bibinfo{year}{2023}.
\newblock \bibinfo{title}{Multiscale structure guided diffusion for image deblurring}, in: \bibinfo{booktitle}{Proceedings of the IEEE/CVF International Conference on Computer Vision}, pp. \bibinfo{pages}{10721--10733}.
\bibitem[{Ren et~al.(2024)Ren, Deng, Zhang, Song, Cao and Yang}]{ren2024fast}
\bibinfo{author}{Ren, W.}, \bibinfo{author}{Deng, S.}, \bibinfo{author}{Zhang, K.}, \bibinfo{author}{Song, F.}, \bibinfo{author}{Cao, X.}, \bibinfo{author}{Yang, M.H.}, \bibinfo{year}{2024}.
\newblock \bibinfo{title}{Fast ultra high-definition video deblurring via multi-scale separable network}.
\newblock \bibinfo{journal}{International Journal of Computer Vision} \bibinfo{volume}{132}, \bibinfo{pages}{1817--1834}.
\bibitem[{Ren et~al.(2017)Ren, Pan, Cao and Yang}]{ren2017videodeblurring}
\bibinfo{author}{Ren, W.}, \bibinfo{author}{Pan, J.}, \bibinfo{author}{Cao, X.}, \bibinfo{author}{Yang, M.H.}, \bibinfo{year}{2017}.
\newblock \bibinfo{title}{Video deblurring via semantic segmentation and pixel-wise non-linear kernel}, in: \bibinfo{booktitle}{Proceedings of the IEEE International Conference on Computer Vision}, pp. \bibinfo{pages}{1077--1085}.
\bibitem[{Richardson(1972)}]{richardson1972bayesian}
\bibinfo{author}{Richardson, W.H.}, \bibinfo{year}{1972}.
\newblock \bibinfo{title}{Bayesian-based iterative method of image restoration}.
\newblock \bibinfo{journal}{JoSA} \bibinfo{volume}{62}, \bibinfo{pages}{55--59}.
\bibitem[{Rota et~al.(2023)Rota, Buzzelli, Bianco and Schettini}]{rota2023video}
\bibinfo{author}{Rota, C.}, \bibinfo{author}{Buzzelli, M.}, \bibinfo{author}{Bianco, S.}, \bibinfo{author}{Schettini, R.}, \bibinfo{year}{2023}.
\newblock \bibinfo{title}{Video restoration based on deep learning: a comprehensive survey}.
\newblock \bibinfo{journal}{Artificial Intelligence Review} \bibinfo{volume}{56}, \bibinfo{pages}{5317--5364}.
\bibitem[{Shafiq and Gu(2022)}]{shafiq2022deep}
\bibinfo{author}{Shafiq, M.}, \bibinfo{author}{Gu, Z.}, \bibinfo{year}{2022}.
\newblock \bibinfo{title}{Deep residual learning for image recognition: A survey}.
\newblock \bibinfo{journal}{Applied Sciences} \bibinfo{volume}{12}, \bibinfo{pages}{8972}.
\bibitem[{Shang et~al.(2025)Shang, Ren, Yang and Zuo}]{weiaggregating2025}
\bibinfo{author}{Shang, W.}, \bibinfo{author}{Ren, D.}, \bibinfo{author}{Yang, Y.}, \bibinfo{author}{Zuo, W.}, \bibinfo{year}{2025}.
\newblock \bibinfo{title}{Aggregating nearest sharp features via hybrid transformers for video deblurring}.
\newblock \bibinfo{journal}{Information Sciences} \bibinfo{volume}{694}, \bibinfo{pages}{121689}.
\bibitem[{Shang et~al.(2021)Shang, Ren, Zou, Ren, Luo and Zuo}]{shang2021bringing}
\bibinfo{author}{Shang, W.}, \bibinfo{author}{Ren, D.}, \bibinfo{author}{Zou, D.}, \bibinfo{author}{Ren, J.S.}, \bibinfo{author}{Luo, P.}, \bibinfo{author}{Zuo, W.}, \bibinfo{year}{2021}.
\newblock \bibinfo{title}{Bringing events into video deblurring with non-consecutively blurry frames}, in: \bibinfo{booktitle}{Proceedings of the IEEE/CVF International Conference on Computer Vision}, pp. \bibinfo{pages}{4531--4540}.
\bibitem[{Shen et~al.(2021)Shen, Cheng, Lu, Zhai, Chen, Asif and Gao}]{shen2021spatial}
\bibinfo{author}{Shen, W.}, \bibinfo{author}{Cheng, M.}, \bibinfo{author}{Lu, G.}, \bibinfo{author}{Zhai, G.}, \bibinfo{author}{Chen, L.}, \bibinfo{author}{Asif, M.S.}, \bibinfo{author}{Gao, Z.}, \bibinfo{year}{2021}.
\newblock \bibinfo{title}{Spatial temporal video enhancement using alternating exposures}.
\newblock \bibinfo{journal}{IEEE Transactions on Circuits and Systems for Video Technology} \bibinfo{volume}{32}, \bibinfo{pages}{4912--4926}.
\bibitem[{Sobel et~al.(1968)Sobel, Feldman et~al.}]{sobel19683x3}
\bibinfo{author}{Sobel, I.}, \bibinfo{author}{Feldman, G.}, et~al., \bibinfo{year}{1968}.
\newblock \bibinfo{title}{A 3x3 isotropic gradient operator for image processing}.
\newblock \bibinfo{journal}{a talk at the Stanford Artificial Project in} \bibinfo{volume}{1968}, \bibinfo{pages}{271--272}.
\bibitem[{Son et~al.(2021)Son, Lee, Lee, Cho and Lee}]{son2021recurrent}
\bibinfo{author}{Son, H.}, \bibinfo{author}{Lee, J.}, \bibinfo{author}{Lee, J.}, \bibinfo{author}{Cho, S.}, \bibinfo{author}{Lee, S.}, \bibinfo{year}{2021}.
\newblock \bibinfo{title}{Recurrent video deblurring with blur-invariant motion estimation and pixel volumes}.
\newblock \bibinfo{journal}{ACM Transactions on Graphics (TOG)} \bibinfo{volume}{40}, \bibinfo{pages}{1--18}.
\bibitem[{Song et~al.(2024)Song, Gai and Da}]{song2024memory}
\bibinfo{author}{Song, G.}, \bibinfo{author}{Gai, S.}, \bibinfo{author}{Da, F.}, \bibinfo{year}{2024}.
\newblock \bibinfo{title}{Memory-based gradient-guided progressive propagation network for video deblurring}.
\newblock \bibinfo{journal}{The Visual Computer} , \bibinfo{pages}{1--16}.
\bibitem[{Su et~al.(2017)Su, Delbracio, Wang, Sapiro, Heidrich and Wang}]{su2017deep}
\bibinfo{author}{Su, S.}, \bibinfo{author}{Delbracio, M.}, \bibinfo{author}{Wang, J.}, \bibinfo{author}{Sapiro, G.}, \bibinfo{author}{Heidrich, W.}, \bibinfo{author}{Wang, O.}, \bibinfo{year}{2017}.
\newblock \bibinfo{title}{Deep video deblurring for hand-held cameras}, in: \bibinfo{booktitle}{Proceedings of the IEEE Conference on Computer Vision and Pattern Recognition}, pp. \bibinfo{pages}{1279--1288}.
\bibitem[{Suin and Rajagopalan(2021)}]{suin2021gated}
\bibinfo{author}{Suin, M.}, \bibinfo{author}{Rajagopalan, A.N.}, \bibinfo{year}{2021}.
\newblock \bibinfo{title}{Gated spatio-temporal attention-guided video deblurring}, in: \bibinfo{booktitle}{Proceedings of the IEEE/CVF Conference on Computer Vision and Pattern Recognition}, pp. \bibinfo{pages}{7802--7811}.
\bibitem[{Sun et~al.(2019)Sun, Zhao, Jiang, Cheng, Xiao, Liu, Mu, Wang, Liu and Wang}]{sun2019high}
\bibinfo{author}{Sun, K.}, \bibinfo{author}{Zhao, Y.}, \bibinfo{author}{Jiang, B.}, \bibinfo{author}{Cheng, T.}, \bibinfo{author}{Xiao, B.}, \bibinfo{author}{Liu, D.}, \bibinfo{author}{Mu, Y.}, \bibinfo{author}{Wang, X.}, \bibinfo{author}{Liu, W.}, \bibinfo{author}{Wang, J.}, \bibinfo{year}{2019}.
\newblock \bibinfo{title}{High-resolution representations for labeling pixels and regions}.
\newblock \bibinfo{journal}{arXiv preprint arXiv:1904.04514} .
\bibitem[{Sun et~al.(2024)Sun, Zhang, Yang, Gao and Fan}]{sun2024long}
\bibinfo{author}{Sun, L.}, \bibinfo{author}{Zhang, J.}, \bibinfo{author}{Yang, Z.}, \bibinfo{author}{Gao, D.}, \bibinfo{author}{Fan, B.}, \bibinfo{year}{2024}.
\newblock \bibinfo{title}{Long-term object tracking based on joint tracking and detection strategy with siamese network}.
\newblock \bibinfo{journal}{Multimedia Systems} \bibinfo{volume}{30}, \bibinfo{pages}{162}.
\bibitem[{Wang et~al.(2021)Wang, Zhang, Jiang, Zhao, Chen and Luo}]{wang2021video}
\bibinfo{author}{Wang, T.}, \bibinfo{author}{Zhang, X.}, \bibinfo{author}{Jiang, R.}, \bibinfo{author}{Zhao, L.}, \bibinfo{author}{Chen, H.}, \bibinfo{author}{Luo, W.}, \bibinfo{year}{2021}.
\newblock \bibinfo{title}{Video deblurring via spatiotemporal pyramid network and adversarial gradient prior}.
\newblock \bibinfo{journal}{Computer Vision and Image Understanding} \bibinfo{volume}{203}, \bibinfo{pages}{103135}.
\bibitem[{Xian et~al.(2024)Xian, Wang and Manocha}]{xian2024mitfas}
\bibinfo{author}{Xian, R.}, \bibinfo{author}{Wang, X.}, \bibinfo{author}{Manocha, D.}, \bibinfo{year}{2024}.
\newblock \bibinfo{title}{Mitfas: Mutual information based temporal feature alignment and sampling for aerial video action recognition}, in: \bibinfo{booktitle}{Proceedings of the IEEE/CVF Winter Conference on Applications of Computer Vision}, pp. \bibinfo{pages}{6625--6634}.
\bibitem[{Xiang et~al.(2020)Xiang, Wei and Pan}]{xiang2020deep}
\bibinfo{author}{Xiang, X.}, \bibinfo{author}{Wei, H.}, \bibinfo{author}{Pan, J.}, \bibinfo{year}{2020}.
\newblock \bibinfo{title}{Deep video deblurring using sharpness features from exemplars}.
\newblock \bibinfo{journal}{IEEE Transactions on Image Processing} \bibinfo{volume}{29}, \bibinfo{pages}{8976--8987}.
\bibitem[{Yan et~al.(2023)Yan, Gong, Wang, Zhang, Zhang and Shi}]{yan2023sharpformer}
\bibinfo{author}{Yan, Q.}, \bibinfo{author}{Gong, D.}, \bibinfo{author}{Wang, P.}, \bibinfo{author}{Zhang, Z.}, \bibinfo{author}{Zhang, Y.}, \bibinfo{author}{Shi, J.Q.}, \bibinfo{year}{2023}.
\newblock \bibinfo{title}{Sharpformer: Learning local feature preserving global representations for image deblurring}.
\newblock \bibinfo{journal}{IEEE Transactions on Image Processing} \bibinfo{volume}{32}, \bibinfo{pages}{2857--2866}.
\bibitem[{Yang et~al.(2020)Yang, Yang, Fu, Lu and Guo}]{yang2020learning}
\bibinfo{author}{Yang, F.}, \bibinfo{author}{Yang, H.}, \bibinfo{author}{Fu, J.}, \bibinfo{author}{Lu, H.}, \bibinfo{author}{Guo, B.}, \bibinfo{year}{2020}.
\newblock \bibinfo{title}{Learning texture transformer network for image super-resolution}, in: \bibinfo{booktitle}{Proceedings of the IEEE/CVF Conference on Computer Vision and Pattern Recognition}, pp. \bibinfo{pages}{5791--5800}.
\bibitem[{Zhang et~al.(2024)Zhang, Xie and Yao}]{zhang2024blur}
\bibinfo{author}{Zhang, H.}, \bibinfo{author}{Xie, H.}, \bibinfo{author}{Yao, H.}, \bibinfo{year}{2024}.
\newblock \bibinfo{title}{Blur-aware spatio-temporal sparse transformer for video deblurring}, in: \bibinfo{booktitle}{Proceedings of the IEEE/CVF Conference on Computer Vision and Pattern Recognition}, pp. \bibinfo{pages}{2673--2681}.
\bibitem[{Zhang et~al.(2022)Zhang, Ren, Luo, Lai, Stenger, Yang and Li}]{zhang2022deep}
\bibinfo{author}{Zhang, K.}, \bibinfo{author}{Ren, W.}, \bibinfo{author}{Luo, W.}, \bibinfo{author}{Lai, W.S.}, \bibinfo{author}{Stenger, B.}, \bibinfo{author}{Yang, M.H.}, \bibinfo{author}{Li, H.}, \bibinfo{year}{2022}.
\newblock \bibinfo{title}{Deep image deblurring: A survey}.
\newblock \bibinfo{journal}{International Journal of Computer Vision} \bibinfo{volume}{130}, \bibinfo{pages}{2103--2130}.
\bibitem[{Zhong et~al.(2020)Zhong, Gao, Zheng and Zheng}]{zhong2020efficient}
\bibinfo{author}{Zhong, Z.}, \bibinfo{author}{Gao, Y.}, \bibinfo{author}{Zheng, Y.}, \bibinfo{author}{Zheng, B.}, \bibinfo{year}{2020}.
\newblock \bibinfo{title}{Efficient spatio-temporal recurrent neural network for video deblurring}, in: \bibinfo{booktitle}{Computer Vision -- ECCV 2020: 16th European Conference, Glasgow, UK, August 23–28, 2020, Proceedings, Part VI}, \bibinfo{publisher}{Springer International Publishing}. pp. \bibinfo{pages}{191--207}.
\bibitem[{Zhu et~al.(2022)Zhu, Xiao, Huang and Zhao}]{zhu2022dastnet}
\bibinfo{author}{Zhu, Q.}, \bibinfo{author}{Xiao, Z.}, \bibinfo{author}{Huang, J.}, \bibinfo{author}{Zhao, F.}, \bibinfo{year}{2022}.
\newblock \bibinfo{title}{Dast-net: Depth-aware spatio-temporal network for video deblurring}, in: \bibinfo{booktitle}{2022 IEEE International Conference on Multimedia and Expo (ICME)}, \bibinfo{organization}{IEEE}. pp. \bibinfo{pages}{1--6}.
\bibitem[{Zhu et~al.(2023)Zhu, Zhou, Zheng, Li, Huang and Zhao}]{zhu2023exploring}
\bibinfo{author}{Zhu, Q.}, \bibinfo{author}{Zhou, M.}, \bibinfo{author}{Zheng, N.}, \bibinfo{author}{Li, C.}, \bibinfo{author}{Huang, J.}, \bibinfo{author}{Zhao, F.}, \bibinfo{year}{2023}.
\newblock \bibinfo{title}{Exploring temporal frequency spectrum in deep video deblurring}, in: \bibinfo{booktitle}{Proceedings of the IEEE/CVF International Conference on Computer Vision}, pp. \bibinfo{pages}{12428--12437}.

\end{thebibliography}

\clearpage
\onecolumn
\begin{center}
    \Large Appendix of ``Video Deblurring by Sharpness Prior Detection and Edge Information''\\
    
\end{center}
\appendix
\section{Autofocus Metrics}\label{app:autofocus}
This section contains a detailed description of each of the $6$ measures utilized for the detection of sharp frames.
Henceforth, given an image $I$, we indicate with $I(\mu,\nu)$ the value of the pixel $(\mu,\nu)$. We indicate the lp-pooling operator with $\Phi^{(p)}_k$, where $p$ is the order of the lp-norm and $k$ is the kernel size. The average pooling is indicated with $\Psi_k$ instead\\
\paragraph{Image Contrast (MIS3)}
First the contrast $C$ is computed by considering 
\begin{equation}
 C(\mu,\nu) = \sum_{i=\mu-1}^{\mu+1} \sum_{j=\nu-1}^{\nu+1} |I(\mu,\nu) - I(i,j)|.
\end{equation}
The metric is then deduced by computing $\Phi^{(1)}_k(C)$ and taking the global average to get a scalar value.
\medskip

\paragraph{Tenengrad Variance (GRA7)}
First the following map is deduced
\begin{equation}
 \phi_{\text{GRA7}} = \Phi^{(2)}_k(\nabla I - \overline{\nabla I})^2 ,
\end{equation}
where $\nabla$ is the Sobel operator, and  \(\overline{\nabla I}\) is the result of $\Psi_k$ with padding, so that the spatial dimensions are compatible. The metric is finally deduced by considering the global average of $\Phi_{\text{GRA7}}$.
\medskip

\paragraph{Energy of Laplacian (LAP1)}
The energy of the Laplacian metric is defined by 
$\phi_{\text{LAP1}}=\Phi_k^{(2)}(\Delta I)^2$, where \(\Delta\) is the Laplacian operator, and then taking the global average.
\medskip

\paragraph{Gray-level Variance (STA3)}
Let $\mu$ be the result of $\Psi_k$ applied to $I$ with padding so that the shape of $I$ is preserved.
The metric is then computed by considering the global average of $\phi_{\text{STA3}}=\Phi_k^{(2)}(I - \mu)^2$ among the pixels.
\medskip

\paragraph{Modified DCT (DCT3)}
The measurement based on the modified discrete cosine transform (DCT) can be efficiently implemented through linear convolution with an $8 \times 8$ mask \(M\), and this measurement is computed over the entire image. The mask \(M\) is defined as:
\(
M =  \begin{bmatrix}
    1 & -1 \\
    -1 & 1 \\
\end{bmatrix} \otimes
\begin{bmatrix}
    1 & 1 \\
    1 & 1 \\
\end{bmatrix}
\)
The transformed image $\phi_{\text{DCT3}} = I \ast M$ is then aggregated by average.
\medskip

\paragraph{Sum of Wavelet Coefficients (WAV1)}
The measurement method is based on the discrete wavelet transform (DWT). Initially, the image is decomposed into four sub-images: \(W_{LH1}\), \(W_{HL1}\), \(W_{HH1}\), and \(W_{LL1}\), representing three detail sub-bands and one coarse approximation sub-band, respectively. The detailed and coarse sub-band information is then used to calculate the focus measurement. The sum of the last three sub-images,
\[
    \phi_{\text{WAV1}} = |W_{LH1}| +  |W_{HL1}| + |W_{HH1}|,
\]
is aggregated by the global sum to deduce the metric. In this work, all wavelet transform coefficients are fixed, thereby avoiding the need to compute the corresponding neighborhood within each sub-band. The terms \(\left|W_{HL1}\right|\) and \(\left|W_{HH1}\right|\) use 1-level DWT with Daubechies-6 filters, while \(\left|W_{LH1}\right|\) uses 2-level DWT with Daubechies-10 filters.





\section{Further methods}
\label{app:not-compared}
This section contains a list of methods and the reasons why we deliberately did not compare them to the \speinet model
\begin{enumerate}
    \item The model proposed in \cite{huang2024effective} has not be tested on the \goproo dataset. Nevertheless, the code is not publicly available.
    \item The model proposed in \cite{kim2024frequency} leverages extra information deduced from an event camera, and hence can not be directly compared to \speinet.
    \item The model proposed in \cite{imani2024stereoscopic} focused on Stereo Blur dataset, and the code is not publicly available.
\end{enumerate}

\section{Video frames of \goprors and \gopros}
In this section, we compared the number of each video sharp frames in \goprors and \gopros. As shown in \Cref{tab:ratios}, The number of videos of \goprors should be less than 50\% of the total number. The number of videos of \gopros should be greater than 50\%.

\begin{table*}[!htbp]
\caption{Actual ratios of the \goprors videos with different ratios $r$, and, in the last column, the measured ratio of the \gopros dataset.}
\centering
\resizebox{\textwidth}{!}{%
\begin{tabular}{lcccccc|c} 
\toprule
\textbf{Video} & \textbf{\goprors(0.02)} & \textbf{\goprors(0.1)} & \textbf{\goprors(0.2)} & \textbf{\goprors(0.3)} & \textbf{\goprors(0.4)} & \textbf{\goprors(0.5)} & \textbf{\gopros} \\ 
\midrule
\textbf{\#1}  & 0.0268 & 0.1081 & 0.2366 & 0.3231 & 0.3475 & 0.5337 & 0.5068 \\ 
\textbf{\#2}  & 0.0093 & 0.1157 & 0.1538 & 0.2868 & 0.4437 & 0.4774 & 0.5478 \\ 
\textbf{\#3}  & 0.0177 & 0.0924 & 0.1849 & 0.3083 & 0.3933 & 0.5476 & 0.5000 \\ 
\textbf{\#4}  & 0.0288 & 0.1488 & 0.1917 & 0.3043 & 0.3475 & 0.4909 & 0.5385 \\ 
\textbf{\#5}  & 0.0280 & 0.1039 & 0.2048 & 0.3164 & 0.3476 & 0.4554 & 0.5074 \\ 
\textbf{\#6}  & 0.0093 & 0.0982 & 0.2320 & 0.3406 & 0.4812 & 0.3819 & 0.4154 \\ 
\textbf{\#7}  & 0.0238 & 0.1279 & 0.3302 & 0.2961 & 0.4454 & 0.5564 & 0.5462 \\ 
\textbf{\#8}  & 0.0381 & 0.0721 & 0.2240 & 0.3191 & 0.4315 & 0.4841 & 0.4722 \\ 
\textbf{\#9}  & 0.0642 & 0.0841 & 0.2756 & 0.3043 & 0.4145 & 0.4812 & 0.4571 \\ 
\textbf{\#10} & 0.0099 & 0.0870 & 0.1429 & 0.2979 & 0.3776 & 0.5119 & 0.5484 \\ 
\textbf{\#11} & 0.0099 & 0.0741 & 0.2791 & 0.2985 & 0.3816 & 0.4938 & 0.5528 \\ 
\midrule
\textbf{Average} & 0.0242 & 0.1011 & 0.2232 & 0.3085 & 0.4010 & 0.4922 & 0.5084 \\ 
\bottomrule 
\end{tabular}}
\label{tab:ratios}
\end{table*}

\section{Different ratios of \speinet}
\Cref{tab:comparison} used \speinet to select $5$ videos from the $11$ videos in the \goprors test set for quantitative representation. 
\begin{table*}[!htbp]
\caption{In each ratio, select the fisrt five classes of \goprors for comparsion. D2Nets are training on \gopros dataset. \speinet(ours) are training on \goprors dataset and other method are training on \goproo dataset.}
\centering
\resizebox{\textwidth}{!}{%
\begin{tabular}{lcccccccccc} 
\toprule 
\textbf{Model} & \multicolumn{2}{c}{\textbf{\#1}} & \multicolumn{2}{c}{\textbf{\#2}} & \multicolumn{2}{c}{\textbf{\#3}} & \multicolumn{2}{c}{\textbf{\#4}} & \multicolumn{2}{c}{\textbf{\#5}} \\ 
\cmidrule(lr){2-3} \cmidrule(lr){4-5} \cmidrule(lr){6-7} \cmidrule(lr){8-9} \cmidrule(lr){10-11}
 & \textbf{PSNR} & \textbf{SSIM} & \textbf{PSNR} & \textbf{SSIM} & \textbf{PSNR} & \textbf{SSIM} & \textbf{PSNR} & \textbf{SSIM} & \textbf{PSNR} & \textbf{SSIM} \\ 
\midrule 
r = 0.02 & 34.503 & 0.9517 & 33.232 & 0.9525 & 34.015 & 0.9483 & 33.905 & 0.9641 & 32.343 & 0.9258\\ 
r = 0.1 &34.447 &0.9524 & 34.229 &0.9591 & 34.541 & 0.9502 & 34.785 & 0.969&33.105  & 0.9333\\ 
r = 0.2 & 35.718 & 0.9603 & 33.525 & 0.956 & 35.038 & 0.9508 &34.685  & 0.9691 &34.166  &0.9383 \\ 
r = 0.3 &36.375 & 0.961 & 34.76 & 0.9609 & 36.235 & 0.9561 & 35.567 & 0.9734 & 34.986 &0.9437 \\ 
r = 0.4 & 36.095 & 0.9623 & 35.515 & 0.9662 & 36.589 & 0.9608 &35.92  &0.9736  &35.251  &0.9484 \\ 
r = 0.5 & 38.192 & 0.9713 & 35.616 & 0.9663 & 38.044 & 0.9653 &36.682  &0.9778  &35.69  &0.9495 \\ 
\bottomrule 
\end{tabular}}
\label{tab:comparison}
\end{table*}

\section{Sharp detectors comparisons}
\label{sec:abl-kernel}
We tested different binary classification models in \Cref{tab:diff-kernel}. Since the F1-score comparison condition is more stringent, we used F1-score for evaluation. Although the prediction results of RandomForest look better, LogistRegression performs best when r=0.2. Combining the F1-score of all ratios, LogistRegression has better stability.
\begin{table*}[!htbp]
\centering
\caption{Training based on GoProRS with different ratios. The shown values are computed on a small validation subset of the training set composed by a 10\% of images.}
\begin{tabular}{lccccccc}
\toprule
\textbf{Model} & \textbf{Indicator} & \textbf{ratio} & \textbf{k=3} & \textbf{k=5} & \textbf{k=7} & \textbf{k=11} & \textbf{k=51} \\
\midrule
LogistRegression & F1-score & 0.2 & 68.7 & 72.9 & 75.2 & 73.3 & 72.1 \\
DecisionTree & F1-score & 0.2 & 57.7 & 58.1 & 59.0 & 62.3 & 60.1 \\
RandomForest   & F1-score & 0.2 & 70.7 & 70.3 & 70.7 & 71.6 & 67.2 \\
\midrule
LogistRegression & F1-score & 0.3 & 70.4 & 71.3 & 71.1 & 74.5 & 74.6 \\
DecisionTree & F1-score & 0.3 & 66.1 & 66.3 & 66.7 & 70.1 & 70.5 \\
RandomForest   & F1-score & 0.3 & 75.9 & 73.9 & 73.8 & 73.9 & 74.0 \\
\midrule
LogistRegression & F1-score & 0.4 & 71.6 & 73.0 & 73.4 & 76.2 & 75.0 \\
DecisionTree & F1-score & 0.4 & 67.5 & 67.9 & 68.2 & 71.5 & 72.0 \\
RandomForest   & F1-score & 0.4 & 78.2 & 75.7 & 75.4 & 75.6 & 76.0 \\
\midrule
LogistRegression & F1-score & 0.5 & 79.3 & 81.1 & 81.4 & 80.2 & 81.8\\
DecisionTree & F1-score & 0.5 & 75.5 & 74.9 & 75.1 & 77.2 & 79.0 \\
RandomForest   & F1-score & 0.5 & 81.4 & 81.7 & 83.9 & 84.9 & 84.4 \\
\bottomrule
\end{tabular}
\label{tab:diff-kernel}
\end{table*}

\end{document}